\documentclass{article}

\usepackage{microtype}
\usepackage{graphicx}
\usepackage{subfigure}
\usepackage{booktabs} 
\usepackage{float}
\usepackage{listings}
\usepackage{rotating}
\usepackage{enumitem}
\usepackage{hyperref}

\usepackage[accepted]{icml2025}

\usepackage{amsmath}
\usepackage{pifont}
\usepackage{amssymb}
\usepackage{mathtools}
\usepackage{amsthm}
\usepackage{colortbl}
\usepackage{booktabs}
\usepackage{makecell}
\usepackage{array}

\usepackage[capitalize,noabbrev]{cleveref}

\theoremstyle{plain}

\theoremstyle{definition}

\theoremstyle{remark}

\usepackage[textsize=tiny]{todonotes}
\usepackage{xcolor}
\usepackage{soul}
\lstdefinestyle{customstyle}{
  basicstyle=\ttfamily\small, 
  keywordstyle=\color{blue}\bfseries, 
  stringstyle=\color{green!40!black}, 
  commentstyle=\color{gray}, 
  frame=single, 
  backgroundcolor=\color{gray!10}, 
  showstringspaces=false, 
  tabsize=2, 
  breaklines=true, 
  morekeywords={challenge,title,description,concepts,difficulty,problem_statement,success,tests_passed,tests_failed,tests_errored,fixed_by_problem_fixer,data_trail,attempt_1,attempt_2}, 
}
\icmltitlerunning{}

\newcommand{\tool}{\textit{Prism}}
\newcommand{\gfouro}{4o}
\newcommand{\gfouromini}{4o-M}
\newcommand{\leightb}{L-8b}
\newcommand{\lseventyb}{L-70b}
\newcommand{\lfourhundredfiveb}{L-405b}

\definecolor{goodgreen}{RGB}{76, 175, 80}
\definecolor{midyellow}{RGB}{255, 235, 59}
\definecolor{badred}{RGB}{244, 67, 54}

\begin{document}

\twocolumn[
\icmltitle{Prism: Dynamic and Flexible Benchmarking of LLMs Code Generation with Monte Carlo Tree Search}



\icmlsetsymbol{equal}{*}

\begin{icmlauthorlist}
\icmlauthor{Vahid Majdinasab}{poly}
\icmlauthor{Amin Nikanjam}{poly,huawei}
\icmlauthor{Foutse Khomh}{poly}
\end{icmlauthorlist}

\icmlaffiliation{poly}{Polytechnique Montreal, Canada}
\icmlaffiliation{huawei}{Huawei Distributed Scheduling and Data Engine Lab, Canada. Work done while at Polytechnique Montreal}

\icmlcorrespondingauthor{Vahid Majdinasab}{vahid.majdinasab@polymtl.ca}

\icmlkeywords{Machine Learning, LLM, Benchmark, MCTS}

\vskip 0.3in
]



\printAffiliationsAndNotice{} 

\begin{abstract}
    The rapid advancement of Large Language Models (LLMs) has outpaced traditional evaluation methods. Static benchmarks fail to capture the depth and breadth of LLM capabilities and eventually become obsolete, while most dynamic approaches either rely too heavily on LLM-based evaluation or remain constrained by predefined test sets. We introduce \tool{}, a flexible, dynamic benchmarking framework designed for comprehensive LLM assessment. \tool{} builds on three key components: (1) a tree-based state representation that models evaluation as a Markov Decision Process, (2) a Monte Carlo Tree Search algorithm adapted to uncover challenging evaluation scenarios, and (3) a multi-agent evaluation pipeline that enables simultaneous assessment of diverse capabilities.
    To ensure robust evaluation, \tool{} integrates structural measurements of tree exploration patterns with performance metrics across difficulty levels, providing detailed diagnostics of error patterns, test coverage, and solution approaches. Through extensive experiments on five state-of-the-art LLMs, we analyze how model architecture and scale influence code generation performance across varying task difficulties. Our results demonstrate \tool{}’s effectiveness as a dynamic benchmark that evolves with model advancements while offering deeper insights into their limitations.        
\end{abstract}

\section{Introduction}\label{sec:introduction}
    The rapid advancement of Large Language Models (LLMs) for code generation has outpaced existing evaluation methodologies. Both static and dynamic benchmarking approaches exhibit fundamental limitations that hinder their ability to comprehensively assess evolving model's capabilities. Static benchmarks, while widely used, quickly become outdated and fail to capture the full scope of an LLM’s reasoning and problem-solving abilities. Dynamic evaluation methods, though designed to be more adaptive, often remain constrained by predefined test sets or rely heavily on LLM-based assessment, introducing biases and inconsistencies. These shortcomings underscore the need for a more robust and adaptable evaluation framework.
    
    Static benchmarks suffer from three key weaknesses. First, they provide only a limited and often superficial assessment of an LLM’s capabilities, reducing evaluation to pass/fail metrics that fail to capture the complexity of the model’s reasoning \cite{mcintosh2024inadequacies, banerjee2024vulnerability, tambon2024assessing}. Second, as benchmarks gain popularity and become evaluation targets, they are increasingly prone to being incorporated into LLMs' training data, leading to data leakage and artificially inflated performance metrics \cite{xu2024benchmark, zhou2023don}. Third, static benchmarks lack flexibility, preventing a targeted evaluation of specific problem-solving strategies or particular aspects of model behavior \cite{mcintosh2024inadequacies}.
    
    To address these limitations, dynamic benchmarking approaches have been introduced \cite{zhu2023judgelm, alpaca_eval, wang2023shepherd, zhang2024darg, li2024treeeval}. These methods typically adopt one of two strategies. The first relies on LLM-as-Judge frameworks, where another LLM is used to evaluate responses \cite{li2024llms}. While this approach offers adaptability, it is inherently unreliable, as evaluation outcomes are constrained by the biases and limitations of the judge model. The second strategy involves dynamically selecting test cases from predefined datasets based on model performance \cite{zhang2024darg, kiela2021dynabench}. Although more structured, these methods remain fundamentally tied to static test sets, limiting their capacity to evolve alongside increasingly capable models.
    
    Given these constraints, a shift in benchmarking methodology is necessary. Instead of relying solely on static or semi-dynamic approaches, evaluation frameworks must be designed to continuously adapt to the evolving capabilities of LLMs. This requires mechanisms that can generate novel and challenging test cases, assess models based on diverse problem-solving strategies, and provide a more fine-grained analysis of strengths and weaknesses.
 
    In this paper, we introduce {\tool{}}, a flexible and dynamic benchmarking framework designed to overcome the limitations of both static and existing dynamic evaluation approaches. Our framework leverages a novel application of Monte Carlo Tree Search (MCTS) \cite{russell2016artificial} to model the evaluation process as a search problem. Specifically, we represent evaluation states using a Markov Decision Process (MDP) \cite{sutton2018reinforcement}, where each state corresponds to a distinct evaluation scenario, and transitions between states reflect increasing levels of complexity. As the search progresses deeper into the tree, the generated evaluation scenarios become progressively more challenging, enabling a structured and adaptive assessment of model capabilities.
    
    The tree-based structure of our MDP allows {\tool{}} to integrate LLM-based agents for targeted analysis tasks at each state while maintaining systematic control over the search process through MCTS-guided exploration. Unlike existing approaches that rely on LLM-as-Judge for direct evaluation, {\tool{}} employs LLM-based agents within well-defined roles, allowing the approach to harness their capabilities without compromising the systematic nature of our search; ensuring that their contributions enhance the benchmarking process without compromising its rigor or consistency. Furthermore, by dynamically generating novel and tailored evaluation scenarios rather than relying on predefined test suites, {\tool{}} enables a more comprehensive assessment of LLMs’ code generation capabilities. Through extensive experimentation with five state-of-the-art LLMs, we demonstrate that {\tool{}} uncovers critical patterns in code generation that traditional benchmarks fail to capture. In particular, our approach provides insights into how LLMs handle increasing complexity and how their performance degrades under challenging coding scenarios, offering a more granular understanding of their strengths and limitations. The primary contributions of our work include:
        \begin{itemize}
            \item A dynamic benchmarking framework that adapts to model capabilities and systematically explores the space of programming challenges.
            \item A novel application of MCTS for structured exploration of code generation capabilities.
            \item A comprehensive multi-agent evaluation architecture that assesses multiple aspects of LLMs' code generation ability.
            \item Detailed empirical analysis of five leading LLMs, revealing new insights about their capabilities and limitations.
        \end{itemize}
    
\section{Related Work}\label{sec:related_work}
    Recent studies have demonstrated how current benchmarking approaches fall short in evaluating current models. Specifically, \cite{xu2024benchmarking}, \cite{roberts2023data}, and \cite{jiang2024investigating} have studied how even careful and extensive attempts to control the training dataset are not capable of preventing data contamination, either because the large size of the training data makes it difficult to filter out undesired data \cite{balloccu2024leak} or because of the lack of diversity in the benchmarks' challenges, alongside models' increasing generalization capabilities \cite{dong2024generalization}. Crowd-sourced benchmarks such as \cite{chiang2024chatbot, open-llm-leaderboard-v2} have proven to be much more effective in evaluating LLMs' capabilities but suffer from 1) a limited coverage of real-world challenges \cite{lin2024wildbench}, and 2) user biases in preference ranking of the responses \cite{chiang2024chatbot}. As such, comprehensive and difficult benchmarks such ARC-AGI \cite{chollet2019measure}, SWE-bench \cite{jimenez2024swebench}, HELM \cite{liang2023holistic}, and HLE \cite{phan2025hle} have been proposed, which include either very difficult problems or data that the models could not have been trained on given the time of their release \cite{white2024livebench, franzmeyer2024hellofresh}. The majority of state-of-the art LLM model providers are currently using these benchmarks. However, with the rapid increase in LLM capabilities, these benchmarks are quickly becoming obsolete, as models achieve increasingly higher scores, rendering the evaluation metrics ineffective in distinguishing their performance~\cite{anthropic2023claude3, jaech2024openai, liu2024deepseek}.
    
    Dynamic benchmarking approaches have been proposed to mitigate these limitations, each addressing one or more of the challenges outlined above. Methods such as \cite{zhu2023judgelm}, \cite{alpaca_eval}, and \cite{wang2023shepherd} adopt an LLM-as-Judge framework, where a more capable LLM—selected based on benchmark performance or fine-tuned for specific evaluation criteria—assesses and ranks the responses of the model under evaluation. Other approaches aim to generate novel and out-of-distribution evaluation scenarios by leveraging LLMs to create new and increasingly complex challenges, as seen in \cite{li2024salad, zhang2024evaluating, zhuge2024agent}. Building on these approaches, frameworks such as \cite{li2024treeeval}, \cite{zhu2024dynamic}, \cite{fan2023nphardeval}, \cite{wang2024benchmark}, and \cite{zhang2024darg} extend dynamic evaluation to provide a more comprehensive assessment of a model's capabilities. Notably, \cite{zhu2024dynamic} employs multiple LLM-based agents to construct fine-grained evaluation scenarios derived from existing benchmarks, ensuring a more precise understanding of model strengths and weaknesses. Meanwhile \cite{zhang2024darg} and \cite{li2024treeeval} represent evaluation processes as graphs or trees, analyzing how the model navigates these structures. However, the stochasticity of LLMs and the variability in their performance introduce challenges in ensuring consistency and reliability in these evaluation methodologies \cite{blackwell2024towards}. Furthermore, offloading the evaluation process to an LLM introduces challenges in the reproducibility and reliability of the evaluation given the judge model's capabilities \cite{thakur2024judging}.

\begin{figure*}[!ht]
    \centering
    \includegraphics[width=0.9\textwidth]{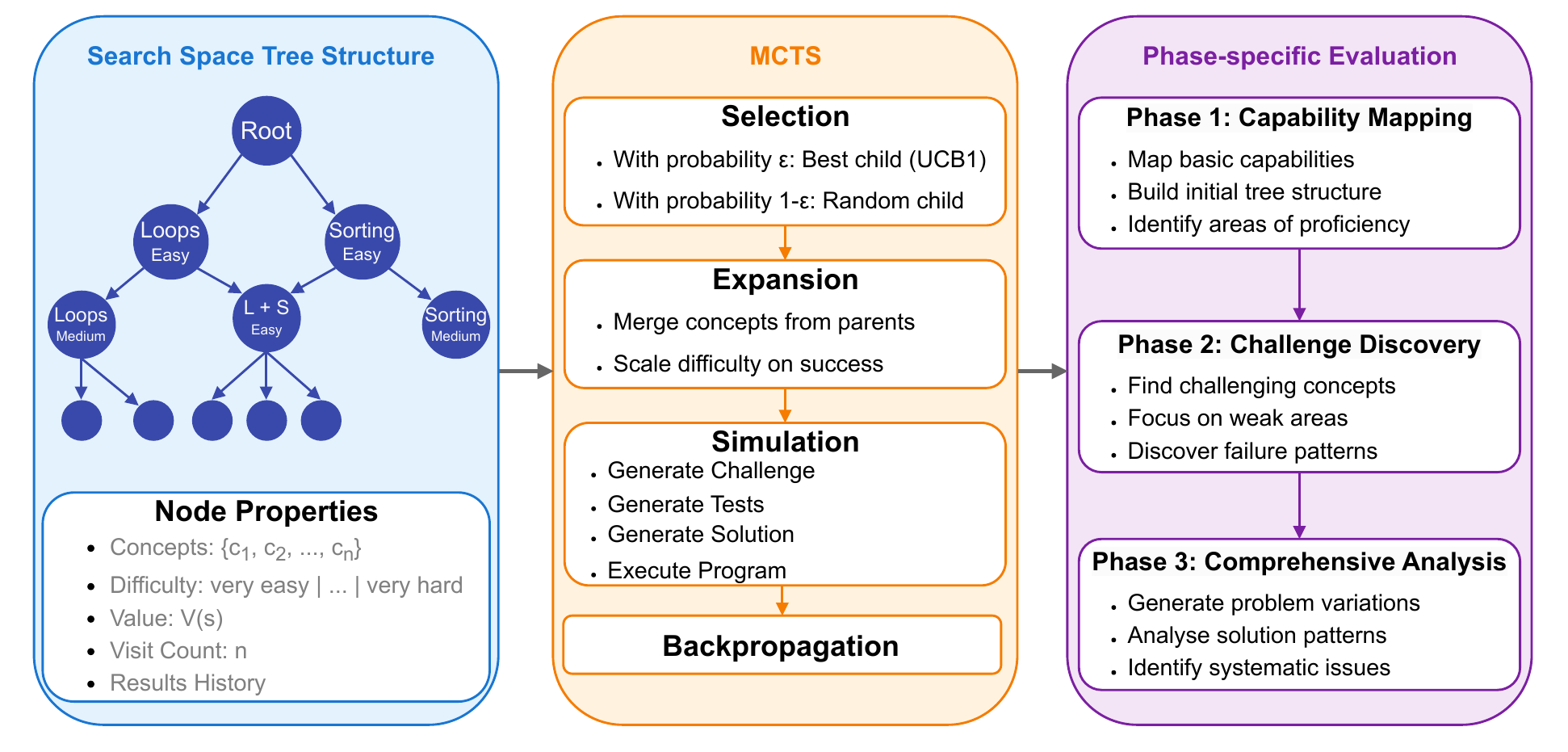}
    \caption{\tool{} is an end-to-end, tree-based, multi-phase evaluation framework for dynamic benchmarking of LLMs across different code generation tasks. It allows for a comprehensive evaluation of the model's capabilities by prioritizing and exploring the search space based on the model's performance using MCTS.}
    \label{fig:prism_methodology}
\end{figure*}
\section{Prism}\label{sec:methodology}  
    \subsection{Overview}\label{subsec:methodology:overview}
        \tool{} is a dynamic benchmarking framework that systematically evaluates LLMs' code generation capabilities through the exploration of coding challenges. We formulate the benchmarking process as a search problem with the objective of identifying model's capabilities and limitations through adaptive exploration of programming concepts with various difficulty levels. 
        
        We model the search space as an MDP, enabling three key properties: (1) Systematic benchmarking through progressively challenging evaluation scenarios, (2) Performance-guided prioritization of under-performing nodes, and (3) Dynamic adaptation of evaluation scenarios based on the model's demonstrated capabilities. In the MDP (search tree), each node represents an evaluation state, encapsulating a unique combination of concepts-difficulty pair. We explore the MDP using $\epsilon$-greedy state selection policies where states are selected using MCTS with a small probability of random state selection to control for LLMs' performance variability. In this manner, we balance the exploration of unexplored areas of the search space with further exploitation of previously explored challenging areas. 
        
        Consider using an LLM to solve a real-world programming problem: for the model to succeed, it must correctly understand the problem, create valid solutions, and effectively resolve issues that are encountered during problem-solving, according to feedback (e.g., raised errors, failing traces, execution issues, etc) \cite{dibia2022aligning}. To evaluate the model's performance, at each node, \tool{} simulates this end-to-end scenario through several steps:
        
        \textbf{Task Comprehension:} The model’s ability to correctly understand the task is the foundation for all subsequent evaluations. A correct understanding ensures the problem is interpreted clearly and precisely \cite{chen2024survey}.\\
        \textbf{Test Creation and Solution Validation:} Once the task is correctly understood, the model must demonstrate the ability to create tests that reflect the task requirements without having access to the solution \cite{zhang2024unseen, li2024large}. Furthermore, it must generate solutions that can pass any valid test, without knowing what the tests are.\\
        \textbf{Error Correction and Feedback Utilization:} If an error is encountered during the process, the model’s understanding of the task,  programming concepts, and syntax, must enable it to fix the problems effectively using structured feedback – including failing test cases (highlighting discrepancies between expected and actual outputs), runtime error diagnostics (e.g., exception types and stack traces), and syntax validation messages from interpreters/compilers. This iterative refinement process tests the model’s capacity to translate error signals into valid corrections \cite{koutcheme2024using}.
        
        This end-to-end process allows \tool{} to explore the search space through a multi-phase evaluation strategy as illustrated in Figure~\ref{fig:prism_methodology}. The process is implemented through a coordinated multi-agent system where \textit{Problem Generator} agents create evaluation scenarios, \textit{Solution Evaluator} agents assess code correctness/quality, and \textit{Pattern Analyzer} agents identify failure patterns - all working collaboratively across phases. The first phase establishes the model’s baseline performance across a variety of concepts and difficulty levels. The second phase targets areas of weakness, probing low-performing nodes to identify systematic gaps in the model's performance. Finally, the third phase generates diverse challenge variations from the low-performing nodes identified in the second phase to identify the root causes of limitations of the model's code generation capabilities.
    
    \subsection{Tree-Based Search}\label{subsec:methodology:tree_based_search_space}
        The search space is formalized as an MDP which is defined by the tuple $(S, A, P, R)$; representing the state space, action space, transition probabilities, and reward function, respectively. The state space $S$ encompasses all possible evaluation scenarios, defined as: 
        \begin{equation}\label{eq:state_space}
            S = \{(c, d) \mid c \subseteq \mathcal{C}, d \in \mathcal{D}\}
        \end{equation}
        where $\mathcal{C}$ represents the set of concepts, and $\mathcal{D}$ defines the difficulty levels. Here, each node in the search tree represents a single state and serves as a single unit of evaluation. The action space $A$ consists of actions for node selection ($A_{select}$) and expansion ($A_{expand}$), allowing both the exploration of existing nodes and the creation of new challenges.
        
        The reward function $R$ provides a quantitative measure of the model's performance at each node. Each phase has a different reward mechanism according to the phase's goal with the reward for each phase being a composite score calculated based on multiple factors including success rates, error penalties, node complexity, and attempt counts, which we discuss in detail in the next Section.

        MCTS relies on Monte Carlo sampling for value estimation. However, given the LLMs' stochastic outputs and the high cost of sampling, we use Temporal Difference (TD) for more efficient value estimation for each node:
        \begin{equation}
            \label{eq:node_value_update}
                v(n) = v_{prev}(n) + \alpha(r - v_{prev}(n))   
        \end{equation}
        where $v(n)$ is the node's value, $r$ is the immediate reward, and $\alpha$ is the hyperparameter that allows tuning the value estimation of each node based on benchmarking requirements.
        
        Our modified MCTS calculates transition probabilities $P$ between states by incorporating visit frequencies and node values. We follow an $\epsilon$- greedy policy to balance between exploration and exploitation:
        \begin{equation}
            \label{eq:traversal_policy}
                \pi(c|n) = \begin{cases}
                    \text{uniform}(children(n)) & 1-\epsilon\\
                    \arg\max_{c \in children(n)} UCB1(c) &  \epsilon 
                \end{cases}
        \end{equation}
        Transitions between nodes (from node $n$ to its child $c$) capture changes in difficulty and the introduction of new concepts, with UCB1 dynamically adjusting transition probabilities between nodes according to the model's historical performance using the node's value $v(n)$ \cite{russell2016artificial}. 
        
        The search process begins by generating foundational nodes that cover basic concepts at the lowest difficulty level. As the search progresses, node creation and expansion are guided by the model's performance, allowing the tree to dynamically adapt to the model's demonstrated capabilities and limitations. This allows \tool{} to adaptively prioritize promising areas while exploring nodes. We present the details of our search approach in Appendix~\ref{sec:appendix:modified_tree_search_algorithm}.
    
    \subsection{Phased Evaluation}\label{subsec:methodology:adaptive_evaluation}
        Our three-phase approach guides MCTS to explore the search space using phase-specific reward functions based on each phase's evaluation strategy: broad capability mapping, targeted challenge discovery, and systematic error analysis, respectively. We include the full state selection policies, reward formulations, and the tree expansion mechanism in \ref{subsec:appendix:modified_tree_search_algorithm:state_selection_policies}, \ref{subsec:appendix:modified_tree_search_algorithm:phase_specific_reward_functions}, and \ref{subsec:appendix:modified_tree_search_algorithm:tree_exapnsion_mechanism}, respectively.
        
        \textbf{Capability Mapping} is the first phase. This phase establishes a baseline assessment of the model’s strengths and weaknesses across a broad concepts-difficulty space. The node scoring mechanism for the first phase focuses on challenge success rate:
        \begin{equation}
        \label{eq:phase_1_reward}
            R_1(s) = b(s) \cdot w(d) + p(s)    
        \end{equation}
        with $b(s)$ being the base score for success, $w(d)$ being the weight for difficulty with harder difficulties having higher weights, and $p(s)$ being the penalty term for failures. In this way, Phase~1 allows for mapping model capabilities: the more successful the model is at solving challenges, the higher the immediate reward, the higher the TD value for the node, and the more MCTS is encouraged to continue exploring the search space to find challenging areas.
        
        \textbf{Challenge Discovery} is the second phase, in which, we leverage the tree constructed in Phase~1 to identify specific areas where the model struggles. By focusing on nodes with low values from Phase~1, this phase continues the search to uncover challenging combinations of concepts and difficulties based on the model's performance. The node scoring mechanism for this phase emphasizes the failure rate and iterative problem-solving efforts:
        \begin{equation}
        \label{eq:phase_2_reward}
            R_2(s) = \lambda(1 - r_{success}) + \gamma \cdot n_{attempts} + \beta \cdot I_{fixer}
        \end{equation}
        with $r_{success}$ being the ratio of successfully passed tests (no errors or failures), $n_{attempts}$ being the number of attempts it took for the model to fix a solution that had failures/errors, and $I_{fixer}$ indicating whether the model required external help to successfully solve the challenge. Here, ($\lambda$, $\gamma$, $\beta$) are hyperparameters that allow for controlling the influence of each term according to benchmarking needs. Using the complement of the success ratio means higher rewards for nodes where the success rate is low. Therefore, nodes that consistently expose the model’s inability to generate correct solutions receive higher rewards and consequently have higher values. This targeted approach produces a set of nodes that indicate the challenging areas of the search space for the model.
        
        \textbf{Comprehensive Evaluation} In this final phase, we conduct an in-depth assessment of the underperforming nodes identified in Phase~2 and use the same reward function as Phase~2. Multiple variations of these nodes are expanded (nodes with the same concept and difficulty but distinct challenge instantiations), enabling differentiation between incidental failures (model struggles with specific scenarios) and systematic limitations (consistent failures across variations). By analyzing performance patterns across these variations, we collect failure traces to investigate surface-level errors and fundamental capability gaps, whether they are results of incorrect syntax, incorrect logic patterns, or incorrect concept implementation. This phase closes the evaluation cycle by revealing not just \textit{where} but \textit{why} the model struggles, providing insights into the root causes of failures. The final tree produced at the end of this phase is presented in Appendix~\ref{sec:appendix:trees}.

    \subsection{Multi-Agent Coordination}\label{subsec:methodology_agents}
        \tool{}'s evaluation strategy is implemented through a coordinated multi-agent system where each agent is responsible for a specific aspect of the benchmarking process across all phases. We define three primary agent types - Problem Generators, Solution Evaluators, and Pattern Analyzers - each with phase-specific roles and coordination patterns:
        
        \textbf{Problem Generators} dynamically create evaluation scenarios (Challenge Designer) and tests (Test Generator) according to each node's concepts-difficulty combinations. \textbf{Solution Evaluators} conduct multi-stage assessments through automated test validation (Test Validator) to investigate whether the tests generated by the model are correct and error analysis (Test Error Analyzer) to investigate the root causes behind failures. \textbf{Pattern Analyzers} perform cross-node analysis (Solution Pattern Analyzer) to detect recurring failure patterns and concept interaction effects through solution and execution trace analysis.

        Agents coordinate through a shared state that tracks solution attempts, test results, and error patterns, enabling adaptive evaluation (e.g., intensifying concept-specific evaluations after repeated failures) based on each node's TD value. The model under benchmark can assume any agent role, allowing targeted capability assessment – for instance, evaluating its problem-solving vs. error-diagnosis skills. We include the full details of our agent implementations, their prompt architectures, and interaction workflows in Appendix~\ref{sec:appendix:agents}.
        
    \subsection{Evaluation Metrics}\label{subsec:methodology_evaluation_metrics }
        We assess LLMs' performance through four metric categories designed to measure different dimensions of the model's code generation capabilities. We include the full definitions and measurement methodologies in Appendix~\ref{sec:appendix:metrics}. Our set of metrics allow for a comprehensive assessment of model capabilities and how models arrive at their solutions:
        
        \textbf{Structural Metrics} focus on the tree and how models perform in the search space. Node counts and depth distributions show where models struggle (persistent exploration) or succeed (rapid convergence), and tree growth patterns demonstrate how challenge complexity impacts performance.\\
        \textbf{Performance Metrics} provide a granular understanding of the model's capabilities using challenge success rates, number of interventions required to fix the model's code, and problem-solving efficiency across concepts and difficulty levels. These metrics identify strengths (high success rate/few attempts) and weaknesses (frequent corrections needed) in each challenge across concepts and difficulties.\\
        \textbf{Mastery Metrics} focus on the model’s progress in understanding and applying concepts over the course of the benchmarking process. These metrics quantify performance stability as challenge complexity increases, success rates on challenges with combinations of concepts, and evolution of solution strategies across benchmarking phases.\\
        \textbf{Diagnostic Metrics} reveal behavioral patterns through solution analysis (preferred coding patterns), error categorization (common failure modes), and test set evaluation (correct tests, testing for corner cases, etc). These metrics capture how the model succeeds or fails in specific scenarios.

\section{Experiments}\label{sec:experiments}      
    \begin{figure*}[h]
        \centering
        \includegraphics[width=0.95\textwidth]{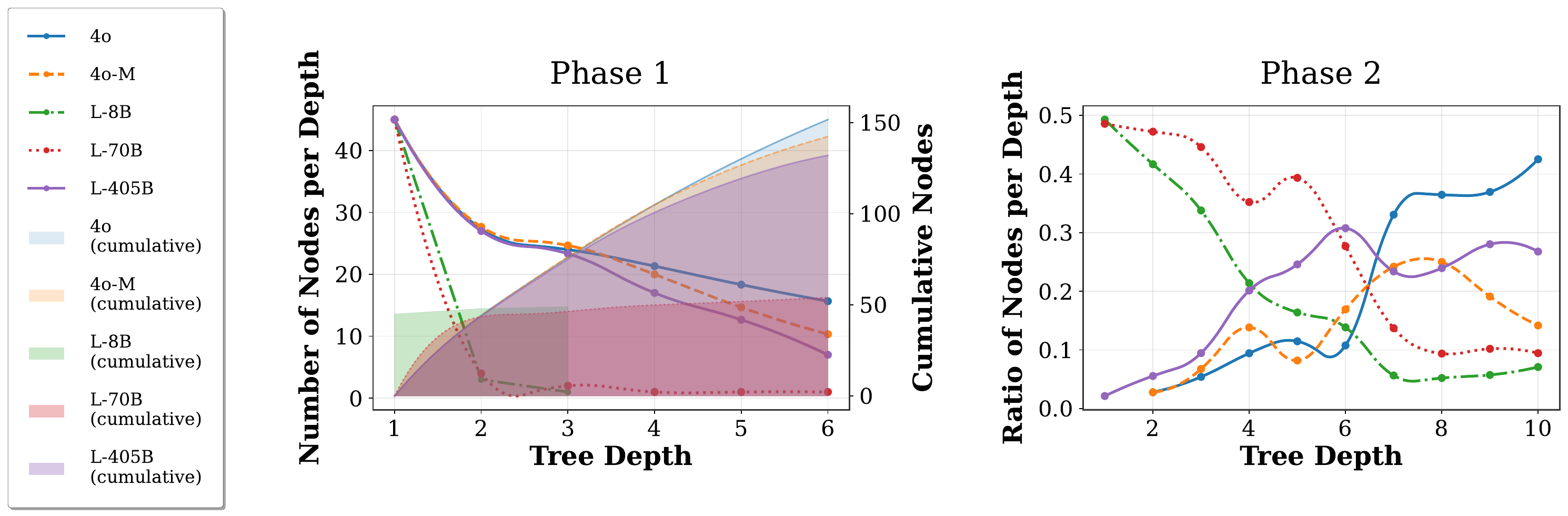}
        \caption{Tree growth analysis across different models. Left panel (Phase~1) shows the node count per tree depth (lines) and the cumulative number of nodes per depth (shaded areas). Right panel (Phase~2) displays the proportion of nodes for each model at each depth in Phase~2, indicating relative search focus across different tree depths.}
        \label{fig:tree_structure}
        \vspace{-1em}
    \end{figure*}
    To show \tool's effectiveness, we evaluate 5 LLMs on their code generation, test creation, and program repair capabilities: GPT4o (\gfouro), GPT4o-mini (\gfouromini) \cite{hurst2024gpt}, LLama3.1-8b (\leightb), LLama3.1-70b (\lseventyb), and Llama3.1-405b (\lfourhundredfiveb) \cite{dubey2024llama}. For our experiments, we use LeetCode (LC)~\cite{leetcode} style programming challenges and {\gfouromini} as the challenge designer to create problems based on specified concepts and difficulty levels. For all LLMs under study, the concepts are chosen similar to the fundamental concepts of computer science in LC with difficulty levels of ``very easy'', ``easy'', ``medium'', ``hard'', and ``very hard'', the same as difficulties of LC challenges. We describe the details of the concepts, concept combinations, and difficulty levels in Appendices~\ref{subsec:appendix:results:concepts} and \ref{subsec:appendix:results:combinations_of_concepts}. The test generation, code generation, and repair tasks are performed by the models under evaluation, while {\gfouro} is used for analyzer agents in Phase~3.  All the reported results are averaged over 3 independent benchmarking runs for all models under study. 

    \subsection{Comparative Analysis}\label{subsec:experiments:comparative_analysis}
        \textbf{Structural Metrics:} 
        Figure~\ref{fig:tree_structure} compares the tree growth and node expansion rates in Phases 1 and 2. As detailed in Section~\ref{subsec:methodology:adaptive_evaluation}, Phase~1's reward function prioritizes task success and difficulty-weighted exploration. Therefore, the Area Under the Curve (AUC) quantifies how effectively models sustain problem-solving capability as challenges become more complex: higher AUCs indicate broader exploration and lower failures. For instance, {\gfouro} achieves 150 nodes in Phase~1, demonstrating robust handling of complex challenges (e.g., multi-concept and high-difficulty tasks), while {\leightb} stalls at 50 nodes, failing beyond basic concepts and easy difficulties (depth$<$4). Phase~2 uses low-scoring Phase~1 nodes to generate targeted challenges, prioritizing task failure and repeated attempts. Therefore, the ratio of generated nodes per depth reveals where models struggle: higher ratios at shallower depths imply difficulty with simpler challenges while increasing ratios at greater depths demonstrate stronger problem-solving capability at complex challenges. We can observe that even though {\gfouromini} has a higher AUC than {\lfourhundredfiveb} in Phase~1 (142 vs 131 nodes), it struggles with complex challenges in Phase~2, while {\lfourhundredfiveb} demonstrates a more consistent exploration of the tree and has a much higher ratio of nodes compared to {\gfouromini} at the end of Phase~2.
        \begin{table}[t]
        \centering
        \caption{Concept Distribution Analysis by Model}
        \label{tab:concept-distribution}
        \resizebox{0.8\linewidth}{!}{%
            \begin{tabular}{lcc|cc}
            \toprule
            & \multicolumn{2}{c}{\gfouro} & \multicolumn{2}{c}{\lfourhundredfiveb} \\
            \cmidrule(lr){2-3} \cmidrule(lr){4-5}
            Concept & Phase~1 & Phase~2 & Phase~1 & Phase~2 \\
            \midrule
            Algorithms & 11.3 & 11.6 & 9.6 & 8.6 \\
            Conditionals & 9.6 & 9.4 & 10.9 & 10.0 \\
            Data Struct. & \cellcolor{yellow}\textbf{9.6} & \cellcolor{yellow}\textbf{11.7} & 10.9 & 9.3 \\
            Dyn. Prog. & 10.3 & 9.7 & \cellcolor{yellow}\textbf{7.3} & \cellcolor{yellow}\textbf{8.0} \\
            Error Hand. & \cellcolor{yellow}\textbf{9.0} & \cellcolor{yellow}\textbf{10.6} & 10.5 & 8.2 \\
            Functions & 10.5 & 8.6 & 10.2 & 11.3 \\
            Loops & 10.9 & 8.2 & \cellcolor{yellow}\textbf{9.7} & \cellcolor{yellow}\textbf{10.8} \\
            Recursion & \textbf{8.9} & 9.2 & \textbf{8.1} & 8.1 \\
            Searching & 9.2 & 9.1 & 12.3 & 13.6 \\
            Sorting & 10.6 & \textbf{11.9} & 10.8 & \textbf{12.1} \\
            \bottomrule
            \end{tabular}
            }
            \vspace{-1em}
        \end{table}
        Table~\ref{tab:concept-distribution} shows the ratio of nodes (percentage) containing a concept per phase for our two most capable models (we include a more detailed analysis in \ref{subsec:appendix:analysis:detailed_results}). For Phase~1, a higher ratio indicates a higher capability to solve challenges with that concept. For Phase~2, however, a higher value indicates struggles with either understanding or solving the challenges with that concept (given each phase's goal and reward function). We observe that both models struggle with "recursion" in Phase~1 and ``sorting" in Phase~2, with {\gfouro} struggling with challenges in ``data structures" and ``error handling" in both phases and {\lfourhundredfiveb} with ``loops" and ``dynamic programming" (low ratio of nodes in Phase~1, high ratio of nodes in Phase~2) as highlighted in the table.
        
        \textbf{Performance Metrics:}
        \begin{table*}[htb]
        \centering
        \caption{Model Capability Analysis by Concept and Difficulty. Values represent failure rates (higher = more challenging). Colors indicate performance: green (good) to red (poor). † indicates primary operational difficulty level (most number of nodes), \checkmark indicates mastered concepts (no failures), and \ding{55} indicates concepts beyond current capability (0 success rate).}
        \label{tab:concept-capabilities}
        \resizebox{0.9\textwidth}{!}{%
            \begin{tabular}{lc|c|c|c|c c c|c|c|c|c c c|c|c|c|c}
            \toprule
                & \multicolumn{5}{c}{Very Easy/Easy} & \multicolumn{1}{c}{} & \multicolumn{5}{c}{Medium} & \multicolumn{1}{c}{} & \multicolumn{5}{c}{Hard/Very Hard} \\
                \cmidrule(lr){2-6} \cmidrule(lr){8-12} \cmidrule(lr){14-18}
                \textbf{Concept} & 4o & 4o-M & L405 & L70 & L8 & & 4o & 4o-M & L405 & L70 & L8 & & 4o & 4o-M & L405 & L70 & L8\\
            \midrule
                Algorithms & \cellcolor{goodgreen!25}\checkmark & \cellcolor{goodgreen!25}\checkmark & \cellcolor{midyellow!56}0.50 & \cellcolor{badred!66}0.81 & \cellcolor{badred!62}0.78† & & \cellcolor{goodgreen!25}\checkmark & \cellcolor{midyellow!61}0.53 & \cellcolor{goodgreen!25}\checkmark & \cellcolor{badred!76}0.91† & \cellcolor{badred!80}0.92 & & \cellcolor{badred!47}0.66† & \cellcolor{badred!76}0.89† & \cellcolor{badred!55}0.73† & \cellcolor{badred!90}\ding{55} & \cellcolor{badred!90}\ding{55}\\[\smallskipamount]
                Conditionals & \cellcolor{goodgreen!25}\checkmark & \cellcolor{goodgreen!25}\checkmark & \cellcolor{goodgreen!25}\checkmark & \cellcolor{badred!65}0.81† & \cellcolor{badred!63}0.79† & & \cellcolor{badred!58}0.75 & \cellcolor{goodgreen!25}\checkmark & \cellcolor{badred!71}0.85 & \cellcolor{badred!76}0.92 & \cellcolor{badred!90}\ding{55} & & \cellcolor{badred!49}0.67† & \cellcolor{badred!74}0.88† & \cellcolor{badred!45}0.64† & \cellcolor{badred!90}\ding{55} & \cellcolor{badred!90}\ding{55}\\[\smallskipamount]
                Data Struct. & \cellcolor{goodgreen!25}\checkmark & \cellcolor{midyellow!70}0.60 & \cellcolor{goodgreen!25}\checkmark & \cellcolor{badred!65}0.80 & \cellcolor{badred!61}0.77† & & \cellcolor{badred!58}0.75 & \cellcolor{midyellow!61}0.53 & \cellcolor{badred!71}0.85 & \cellcolor{badred!76}0.98† & \cellcolor{badred!90}\ding{55} & & \cellcolor{badred!49}0.67† & \cellcolor{badred!75}0.88† & \cellcolor{badred!50}0.68† & \cellcolor{badred!90}\ding{55} & \cellcolor{badred!90}\ding{55}\\[\smallskipamount]
                Dyn. Prog. & \cellcolor{goodgreen!25}\checkmark & \cellcolor{midyellow!70}0.60 & \cellcolor{midyellow!56}0.50 & \cellcolor{badred!61}0.77† & \cellcolor{badred!62}0.78† & & \cellcolor{goodgreen!25}\checkmark & \cellcolor{midyellow!61}0.53 & \cellcolor{badred!71}0.85 & \cellcolor{badred!76}0.88 & \cellcolor{badred!90}\ding{55} & & \cellcolor{badred!48}0.67† & \cellcolor{badred!76}0.89† & \cellcolor{badred!53}0.71† & \cellcolor{badred!90}\ding{55} & \cellcolor{badred!90}\ding{55}\\[\smallskipamount]
                Error Hand. & \cellcolor{goodgreen!25}\checkmark & \cellcolor{goodgreen!25}\checkmark & \cellcolor{goodgreen!25}\checkmark & \cellcolor{badred!67}0.82† & \cellcolor{badred!62}0.78† & & \cellcolor{badred!58}0.75 & \cellcolor{midyellow!61}0.53 & \cellcolor{badred!71}0.85 & \cellcolor{badred!90}\ding{55} & \cellcolor{badred!90}\ding{55} & & \cellcolor{badred!48}0.67† & \cellcolor{badred!76}0.89† & \cellcolor{badred!53}0.71† & \cellcolor{badred!90}\ding{55} & \cellcolor{badred!90}\ding{55}\\[\smallskipamount]
                Functions & 
                \cellcolor{goodgreen!25}\checkmark & \cellcolor{midyellow!70}0.60 & \cellcolor{goodgreen!25}\checkmark & \cellcolor{badred!66}0.82† & \cellcolor{badred!63}0.79† & & \cellcolor{badred!58}0.75 & \cellcolor{midyellow!61}0.53 & \cellcolor{badred!71}0.85 & \cellcolor{badred!90}\ding{55} & \cellcolor{badred!76}0.92 & & \cellcolor{badred!47}0.66† & \cellcolor{badred!77}0.90† & \cellcolor{badred!52}0.70† & \cellcolor{badred!90}\ding{55} & \cellcolor{badred!90}\ding{55}\\[\smallskipamount]
                Loops & 
                \cellcolor{goodgreen!25}\checkmark & \cellcolor{goodgreen!25}\checkmark & \cellcolor{goodgreen!25}\checkmark & \cellcolor{badred!66}0.81 & \cellcolor{badred!63}0.79† & & \cellcolor{goodgreen!25}\checkmark & \cellcolor{midyellow!61}0.53 & \cellcolor{badred!71}0.85 & \cellcolor{badred!76}0.81† & \cellcolor{badred!80}0.92 & & \cellcolor{badred!49}0.67† & \cellcolor{badred!75}0.88† & \cellcolor{badred!53}0.71† & \cellcolor{badred!90}\ding{55} & \cellcolor{badred!90}\ding{55}\\[\smallskipamount]
                Recursion &
                \cellcolor{goodgreen!25}\checkmark & \cellcolor{goodgreen!25}\checkmark & \cellcolor{midyellow!56}0.50 & \cellcolor{badred!63}0.79† & \cellcolor{badred!61}0.77† & & \cellcolor{goodgreen!25}\checkmark & \cellcolor{midyellow!61}0.53 & \cellcolor{badred!71}0.85 & \cellcolor{badred!90}\ding{55} & \cellcolor{badred!90}\ding{55} & & \cellcolor{badred!47}0.66† & \cellcolor{badred!76}0.89† & \cellcolor{badred!49}0.68† & \cellcolor{badred!90}\ding{55} & \cellcolor{badred!90}\ding{55}\\[\smallskipamount]
                Searching &
                \cellcolor{goodgreen!25}\checkmark & \cellcolor{goodgreen!25}\checkmark & \cellcolor{midyellow!56}0.50 & \cellcolor{badred!65}0.80† & \cellcolor{badred!62}0.78† & & \cellcolor{badred!58}0.75 & \cellcolor{midyellow!61}0.53 & \cellcolor{badred!71}0.85 & \cellcolor{badred!90}\ding{55} & \cellcolor{badred!90}\ding{55} & & \cellcolor{badred!48}0.67† & \cellcolor{badred!76}0.89† & \cellcolor{badred!52}0.70† & \cellcolor{badred!90}\ding{55} & \cellcolor{badred!90}\ding{55}\\[\smallskipamount]
                Sorting &
                \cellcolor{goodgreen!25}\checkmark & \cellcolor{midyellow!70}0.60 & \cellcolor{goodgreen!25}\checkmark & \cellcolor{badred!64}0.79† & \cellcolor{badred!62}0.78† & & \cellcolor{goodgreen!25}\checkmark & \cellcolor{midyellow!61}0.53 & \cellcolor{badred!71}0.85 & \cellcolor{badred!76}0.92 & \cellcolor{badred!80}0.98 & & \cellcolor{badred!48}0.66† & \cellcolor{badred!76}0.89† & \cellcolor{badred!53}0.71† & \cellcolor{badred!90}\ding{55} & \cellcolor{badred!90}\ding{55}\\
            \bottomrule
            \end{tabular}
            }
            \vspace{-1em}
        \end{table*}
        Table~\ref{tab:concept-capabilities} displays the capability analysis at the end of the benchmark, with values representing \emph{failure rates} across concepts and difficulty levels. The primary operational capability of a model for each concept is determined by the ratio of nodes (concept-difficulty pairs) explored in the search tree and their average failure rates across 3 independent runs. \tool{} dynamically explores the search space to find challenging areas for the model under evaluation, and once these areas are determined, they are explored iteratively to determine the root causes behind these failures. We can observe that {\gfouro} has no failures in all easy-level challenges, demonstrating strong capability in basic programming challenges. However, its performance degrades at medium/hard difficulties, particularly with ``dynamic programming" and ``data structures", indicating limitations in handling programming challenges that require in-depth reasoning. Unlike {\gfouro}, {\lfourhundredfiveb} fails in some easy challenges, but shows lower failure rates on easy/medium challenges compared to the rest of the models (barring \gfouro). Nonetheless, similar to {\gfouro}, it struggles with hard/very hard challenges that require reasoning and integration of multiple concepts, such as ``dynamic programming", ``algorithms" and ``functions". {\gfouromini} has higher overall failure rates between the 3 top models, especially in challenges requiring compositional reasoning such as ``loops", ``functions", ``conditionals", and ``recursion" indicating struggles with maintaining coherence and correctness in more complex coding challenges. These failures are raised when these concepts are combined (e.g., challenges with both loops and conditional concepts) as shown in Figure~\ref{fig:4o_mini_concept_heatmap} which we explain in detail in Appendix~\ref{subsec:appendix:detailed_analysis:effects_of_scale}. The smaller models, {\lseventyb} and {\leightb}, exhibit distinct performance profiles: {\lseventyb} struggles with challenges with easy difficulties and shows increased failure rates on medium difficulty challenges, indicating a limited capacity for complex tasks. {\leightb} shows high failure rates across all difficulty levels and concepts, highlighting fundamental limitations in its code generation capabilities. These areas directly tie into concept mastery, which we discuss in the rest of this section, and provide more detailed analyses of models' performance and the effects of model scale in Appendices~\ref{subsec:appendix:analysis:detailed_results} and \ref{subsec:appendix:detailed_analysis:effects_of_scale}.
        
        \textbf{Mastery Metrics:} 
        \begin{figure*}[h]
            \centering
            \includegraphics[width=\textwidth]{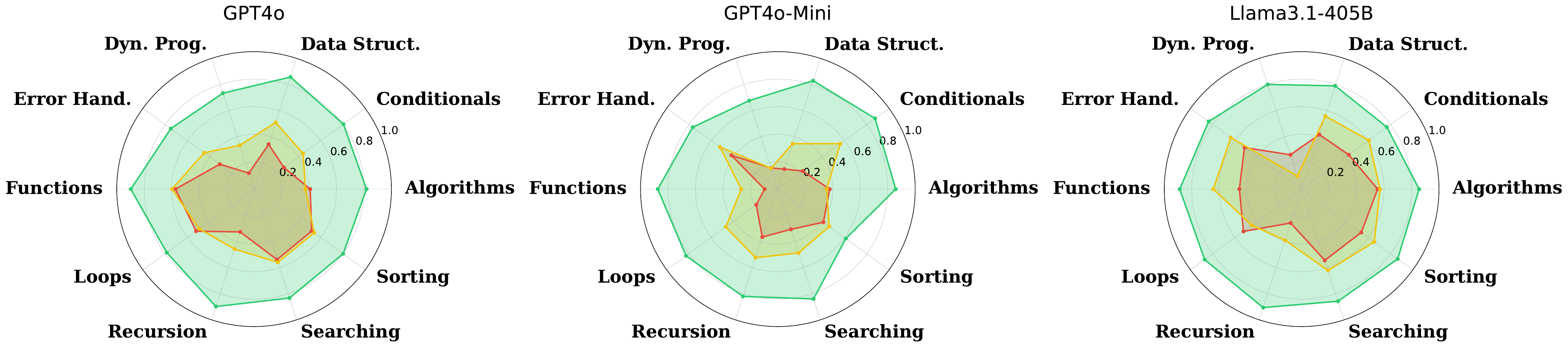}
            \caption{Radar plots showing the performance of {\gfouro}, {\gfouromini}, and {\lfourhundredfiveb} across concepts per each difficulty level. Green: (very easy/easy), Yellow (medium), Red: (hard/very hard). The radial axis represents the success rate (between 0 and 1), while the circumferential axis shows different programming concepts. Higher values indicate better performance.}
            \label{fig:concept_radar}
        \end{figure*}
        Figure~\ref{fig:concept_radar} shows the \emph{success rates} per concept and difficulty level across Phases 1 and 2 for our top models, detailing how each model performs in specific programming concepts and at difficulty levels. While performance metrics focus on overall model performance (i.e., success/failure in solving challenges), mastery metrics highlight exactly which concepts the models handle well and where they struggle. We can observe that for ``dynamic programming", ``recursion", and ``data structures" (concepts that require reasoning and careful solution design), all three models consistently fail compared to other concepts regardless of the difficulty level. Specifically, while {\gfouro} demonstrates better performance than other models on medium-difficulty challenges, its performance rapidly degrades once the challenges require complex conditional logic or advanced dynamic programming strategies. Meanwhile, as observed in Table~\ref{tab:concept-capabilities}, {\gfouromini} struggles with compositional reasoning: challenges that require advanced ``algorithms", ``data structures", and ``dynamic programming". Such failures are encountered in challenges where multiple function calls, multiple/nested conditional branches, or multi-level recursion are required. Also, {\gfouromini} has lower success rates compared to {\gfouro} and {\lfourhundredfiveb}, suggesting a limited capacity to handle complex challenges overall. On the other hand, {\lfourhundredfiveb} demonstrates a strong performance on easier challenges and a moderate performance on medium-difficulty challenges, especially when the challenges center on well-defined loops or basic data structure handling. However, its success rates drop significantly for more complex challenges requiring in-depth reasoning or the integration of multiple concepts (e.g., combining ``dynamic programming" with ``functions"). We include a more detailed analysis of concept combinations and their effects on model performance in Appendix~\ref{subsec:appendix:analysis:detailed_results}. By mapping success rates to each concept-difficulty pair, concept mastery metrics pinpoint common failure modes and help distinguish the models’ capabilities. For example, if a model consistently solves easy ``loop" challenges but shows lower success rates on ``nested loops", this highlights specific weaknesses in handling layered control flow. The challenging areas identified in Phase~1 and explored in Phase~2, show \textit{where} the models fail. Phase~3 builds upon these areas to investigate \textit{how} the models fail.

        \textbf{Diagnostic Metrics:}
        \begin{figure}[!htb]
            \centering
            \includegraphics[width=\linewidth,height=11cm, keepaspectratio]{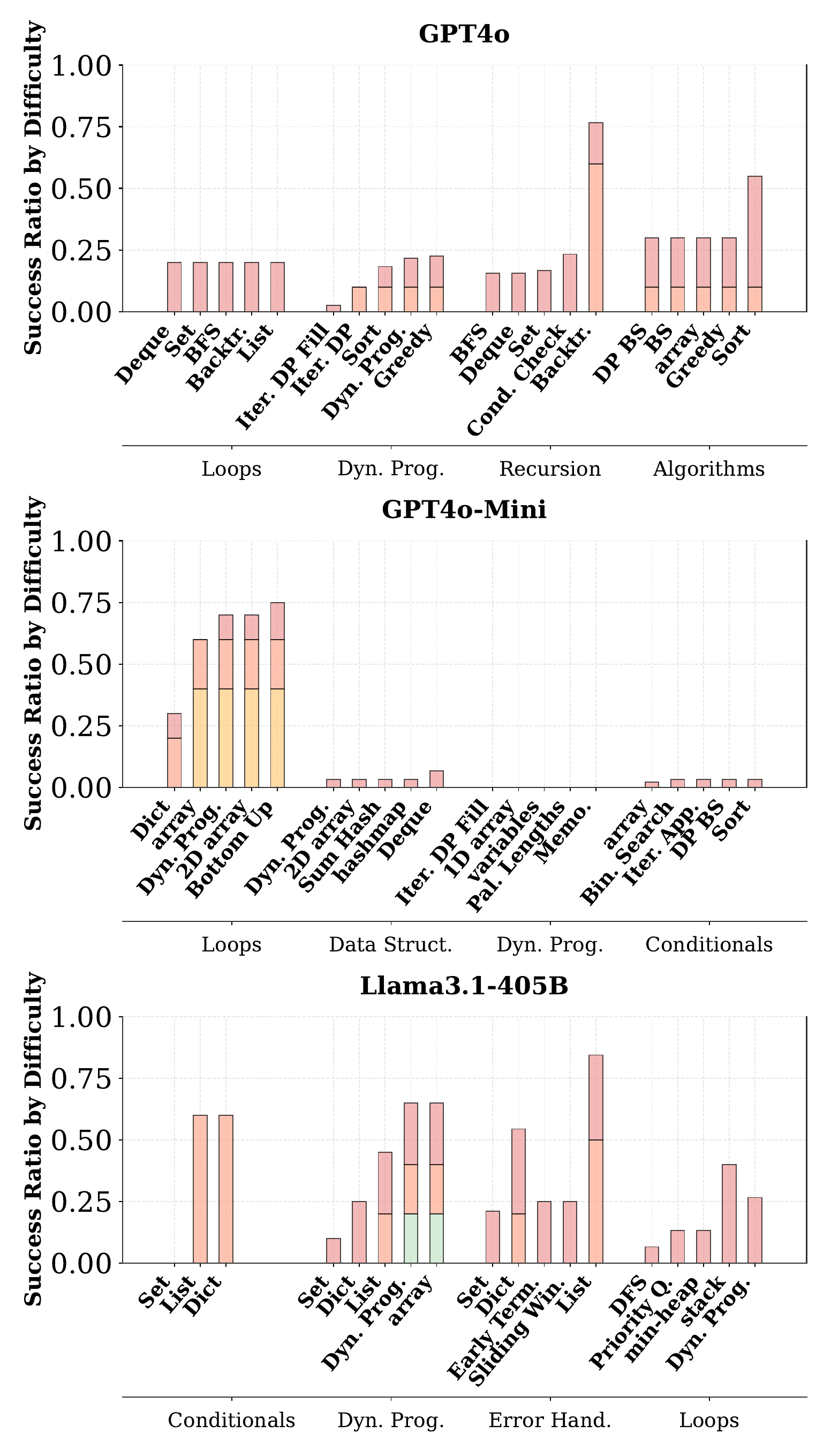}
            \caption{Success ratios for the most challenging programming patterns, grouped by the four most challenging concepts for each model, for {\gfouro}, {\gfouromini}, and {\lfourhundredfiveb}. Stacked bars represent performance across difficulty levels. Higher stacks indicate better overall performance. Results highlight model-specific weaknesses in handling complex programming concepts and patterns. Green (very easy/easy), Yellow: (medium), Red: (hard/very hard)}
            \label{fig:pattern_concept}
            \vspace{-1.5em}
        \end{figure}
        Figure~\ref{fig:pattern_concept} shows the success ratios of the top-performing models for the four highest failure rate concepts from Table~\ref{tab:concept-capabilities}, grouped by the top five programming patterns found in the solutions of each model. Notably, {\gfouro} struggles significantly with ``dynamic programming", even when the concept is not explicitly in the challenge. On the other hand, {\gfouromini} consistently fails in challenges involving composite problems (combinations of multiple concepts), ``complex data structures", or ``dynamic programming", regardless of how it attempts to solve the challenge. Finally, {\lfourhundredfiveb} shows the lowest success ratios for simple ``data structures" and ``tree/graph traversal". In contrast to the other top models, our analysis shows that {\lfourhundredfiveb}'s failures are not due to a lack of understanding of the problem itself; instead, they come from failures in following instructions and programming syntax as indicated by the presence of built-in data types (set, list, dict, etc.) in the failed solutions. Analyzing the failing nodes reveals that while the logic and pseudocode are often correct, {\lfourhundredfiveb} frequently makes errors such as hallucinating keys in built-in types, misplacing code snippets, or failing to follow the system prompt's format, which lead to immediate rejection of solutions by the framework. Comprehensive comparisons of our analyses across models are discussed in Appendix \ref{sec:appendix:results}.

        As shown above, unlike static benchmarks, {\tool{}} dynamically evaluates the LLMs' capabilities. Using agents allows for comprehensive analysis of failures and their root causes, enabling granular diagnostics that static benchmarks do not provide: \textit{where} models struggle and \textit{how} they fail. By modeling the search space as an MDP, using MCTS, and iterative phase transitions, {\tool{}} adaptively prioritizes challenging areas unique to each model, allowing precise, and systematic evaluation that prior dynamic frameworks lack. 

\section{Conclusion and Limitations}\label{sec:conclusion}
    In this paper, we introduced \tool{}, which models the benchmarking search space as an MDP and uses MCTS to explore this space, allowing for evaluating model capabilities in a manner that traditional benchmarks miss. Unlike prior approaches, \tool{} dynamically analyzes how models approach problems, adapt to feedback, and handle increasing complexity. While our examples focus on code generation, \tool{} is adaptable to other domains through customizable agents and scoring adjustments. We recognize two limitations to \tool{} as it is presented: First, \tool{}'s automated evaluation depends on judge models (GPT-4o in our implementation), even though the judge models can be changed to any available model, there still exists the capability ceilings for analyzing highly complex outputs for Phase~3 analysis. Second, the stochasticity of LLMs introduces variability in results across runs, requiring multiple executions to establish statistical reliability. As we have explained throughout the paper and in detail in the appendix, we manage this variability at each step.
    However, detailed metrics may differ between trials.

\newpage
\section*{Impact Statement}
    This paper presents {\tool{}}, a novel framework for evaluating the code generation capabilities of LLMs. {\tool{}} introduces a new methodology for dynamically assessing these capabilities, providing a foundation for understanding and improving LLM-generated code. The growing adoption of LLMs in software development, along with continued advancements in their capabilities, underscores the need for benchmarking approaches that can keep pace with these developments. To address this, our work systematically identifies failure cases and limitations in LLM-generated code, helping to improve model safety and reliability. By enabling the early detection of potential risks—such as security vulnerabilities, unintended biases, and incorrect outputs—our framework contributes to the development of more robust and trustworthy LLMs for code generation. Given the broader implications of our work, there are many potential societal consequences. However, as our focus is solely on evaluating models' code-generation capabilities, we do not feel any of these consequences need to be specifically highlighted here.
    
    
    

\nocite{langley00}

\bibliography{example_paper}
\bibliographystyle{icml2025}

\newpage
\appendix
\onecolumn

\section{Modified Tree Search Algorithm}\label{sec:appendix:modified_tree_search_algorithm}
    We formulate the search tree as an MDP $\mathcal{M} = (\mathcal{S}, \mathcal{A}, \mathcal{P}, \mathcal{R})$ where:
    \begin{itemize}
        \item $\mathcal{S}$ represents the state space of all possible concept and difficulty combinations,
        \item $\mathcal{A}$ defines the action space including concept combination and difficulty adjustment,
        \item $\mathcal{P}: \mathcal{S} \times \mathcal{S} \rightarrow [0,1]$ captures transition dynamics between states,
        \item $\mathcal{R}: \mathcal{S} \times \mathcal{A} \rightarrow \mathbb{R}$ defines the phase-specific reward functions.
    \end{itemize}
    
    We modify MCTS to efficiently explore this MDP through: (1) Temporal Difference (TD) for robust value estimation in stochastic environments (i.e., node scoring), (2) a multi-parent tree structure allowing for concept combinations, and (3) phase-specific policies optimizing exploration-exploitation balance. Below, we detail how the tree is constructed and traversed.

    \subsection{Value Updates and Tree Structure}\label{subsec:appendix:modified_tree_search_algorithm:value_updates_and_tree_structure}
        \subsubsection{TD Scoring for Nodes}
            As mentioned in Section~\ref{subsec:methodology:tree_based_search_space}, MCTS relies on Monte Carlo sampling for value estimation. However, given the stochastic nature of LLMs' outputs and the high cost of sampling, we use TD for more efficient value updates as described in Equation~\ref{eq:node_value_update}:
            \begin{equation*}
                v(n) = v_{prev}(n) + \alpha(r - v_{prev}(n))    
            \end{equation*}
            where $v(n)$ is the node's value, $r$ is the immediate reward, and $\alpha$ is the hyperparameter that allows tuning the value estimation of each node based on benchmarking requirements. 
        
        \subsubsection{Multi-Parent Tree Structure}
            In \tool{}, each node in the search tree can have multiple parent nodes instead of one as each parent represents a unique combination of concepts and difficulties. To capture concept combinations effectively, we extend MCTS's UCB1 \cite{russell2016artificial} to support multi-parent nodes:
            \begin{equation}
            \label{eq:ucb1}
                UCB1(n) = \frac{v(n)}{N(n)} + C\sqrt{\frac{\ln(\sum_{i \in parents(n)} N(p))}{N(n)}}
            \end{equation}
            with $v(n)$ being the node $n$'s value, $N(n)$ being the number of times $n$ has been visited, $C$ being the exploration constant, and $parents(n)$ represents the set of parents. In this manner, a failure to solve the challenges at one node is indicative of the model showing a behavior of interest for each concept and difficulty level in that node.
        
    \subsection{State Selection Policies}\label{subsec:appendix:modified_tree_search_algorithm:state_selection_policies}
        Studies have shown that controlling LLMs' stochasticity through low-temperature settings (e.g., T $\approx$ 0) systematically reduces the diversity of their outputs \cite{renze2024effect,peeperkorn2024temperature}. Although this performance degradation may be minor in some contexts, it becomes crucial when the objective is to comprehensively benchmark a model’s capabilities. As shown by \cite{xu2022systematic,ouyang2023llm}, for code generation, at higher temperatures, LLMs explore novel solutions more effectively, while at near-zero temperatures, outputs become repetitive and risk underestimating true performance boundaries. In \tool{}, while we provide temperature as a tunable parameter, we preserve recommended temperature ranges rather than enforcing low-temperature values. This ensures thorough exploration of the search space but also introduces the problem of stochasticity in LLMs' responses and subsequent performance variations as a result. As mentioned in Section~\ref{sec:related_work}, these performance variations are especially problematic in dynamic benchmarks that rely on LLMs as judges, where even small changes in output can lead to different assessment results. As such, we need to consider that the score for a node might not be representative of the LLM's true capability due to performance variations. To address this problem without the risk of underestimating LLMs' performance by setting low-temperature values, we define $\epsilon$-greedy state selection policies to traverse the tree as described in Equation~\ref{eq:traversal_policy}. These policies mitigate the drawbacks of purely deterministic approaches (e.g., solely using UCB1 for traversal or setting the temperature to zero) by balancing exploration and exploitation and ensuring comprehensive capability assessment while mitigating performance variations. We detail the policies for each phase below.

        \subsubsection{Phase~1: Capability Mapping}\label{subsubsec:appendix:modified_tree_search_algorithm:state_selection_policies:phase_1}
            At the very beginning of the evaluation, the root generates multiple nodes as starting points for the search process. However, since the search requires the nodes to be evaluated first, the policy for evaluating the root's children (initial nodes) is defined as:
            \begin{equation}
            \label{eq:phase_1_initial_policy}
                \pi_1^{root}(n) = \begin{cases}
                    \frac{\sum_{n' \in N} {v(n')}}{v(n) + \mu} & \text{if } \exists n': v(n') > 0 \\
                    \frac{1}{|N|} & \text{otherwise}
                \end{cases}
            \end{equation}
            where $\mu$ prevents division by zero and the inverse of the node's value encourages early exploration of the unvisited nodes. This policy allows for the exploration of the initial nodes to establish a starting point for the search.
            Once initial evaluations are complete, we use an $\epsilon$-greedy policy for traversing the tree:
            \begin{equation}
            \label{eq:phase_1_traversal_policy}
                \pi_1^{traverse}(c|n) = \begin{cases}
                    \text{uniform}(children(n)) & \text{with probability } 1-\epsilon_1\\
                    \arg\max_{c \in children(n)} UCB1(c) & \text{with probability } \epsilon_1 
                \end{cases}
            \end{equation}
            This policy accounts for the stochasticity of LLMs' responses by using a uniform exploration component to mitigate the impact of occasional LLM performance variations. This ensures a thorough exploration of the LLM's capability while still focusing on promising directions using MCTS.
        
        \subsubsection{Phase~2: Challenge Discovery}\label{subsubsec:appendix:modified_tree_search_algorithm:state_selection_policies:phase_2}
            Similar to Phase~1, Phase~2 uses an $\epsilon$-greedy policy that focuses on challenging scenarios while maintaining exploration to mitigate LLMs' performance variations. Therefore, even though our search is guided by UCB1, we consider a small probability of selecting another state in our policy:
            \begin{equation}
            \label{eq:phase_2_traversal_policy}
                \pi_2(c|n) = \begin{cases}
                    \text{uniform}(children(n)) & \text{with probability }  1-\epsilon_2 \\
                    \operatorname*{arg\,max}_{c \in children(n)} UCB1(c) & \text{with probability } \epsilon_2 
                \end{cases}
            \end{equation}
        
        For both Phases 1 and 2, $\epsilon_1$ and $\epsilon_2$ can be tuned based on the benchmarking requirements, with higher values resulting in more stochastic state selections and enable more exploration of the search space regardless of how the model under benchmark performs.
        
        \subsubsection{Phase~3: Comprehensive Evaluation}\label{subsubsec:appendix:modified_tree_search_algorithm:state_selection_policies:phase_3}
            In Phase~3, state selection becomes deterministic based on node value thresholds (i.e., which nodes to select from Phase~2) which can be tuned depending on the benchmark's desired difficulty level:
            \begin{equation}
            \label{eq:phase_3_traversal_policy}
                \pi_3(c|n) = \begin{cases}
                    1 & \text{if } v(c(n)) > \theta_c \\
                    0 & \text{otherwise}
                \end{cases}
            \end{equation}
            with $\theta_c$ being the TD value threshold. Lower thresholds mean that more nodes are selected for analysis (lower TD values) in Phase~3 and this therefore increases the benchmark's overall analysis granularity.
    
    \subsection{Phase-Specific Reward Functions}\label{subsec:appendix:modified_tree_search_algorithm:phase_specific_reward_functions}
        As mentioned in Section~\ref{subsec:methodology:adaptive_evaluation}, in each phase we employ different mechanisms to calculate the immediate reward received. We describe the details of each phase's reward calculation in the following.
    
        \subsubsection{Phase~1: Capability Mapping}\label{subsubsec:appendix:modified_tree_search_algorithm:phase_specific_reward_functions:phase_1}
            In Phase~1, the goal is to thoroughly map the capabilities of the model under study. Therefore, the reward function is focused on task success: the better the model is at successfully passing a challenge at a node, the higher the reward that it will receive. The reward function for this phase is defined in Equation~\ref{eq:phase_1_reward}:
            \begin{equation*}
                R_1(s) = b(s) \cdot w(d) + p
            \end{equation*}
            with $b(s)$ representing the base reward given to the model if it can pass the challenge regardless of the challenge's complexity, or the number of attempts it took the model to solve it. $w(d)$ is the weight assigned to each difficulty level with higher difficulty levels having a higher weight. This way, the more challenging the problem the model has solved, the higher the reward it receives. Finally, the performance penalty, $p$, is defined as:
            \begin{equation}
            \label{eq:reward_penalty_formula}
                p = (r_{failed} \cdot P_{failure}) + (r_{errors} \cdot P_{error}) + ((n_{attempts}-1) \cdot P_{attempt}) + P_{fixer} \cdot I_{fixer}
            \end{equation}
            with $r_{failed}$ being the ratio of tests the model's solution failed, $r_{errors}$ being the ratio of errors in the model's solution, $n_{attempts}$ being the number of attempts it took for the model to solve the challenge, and $I_{fixer}$ being 1 if the \textit{Fixer} agent was required to fix the model's solution and 0 otherwise. $P_{failure}$, $P_{error}$, $P_{attempt}$, and $P_{fixer}$ are the weights assigned to each penalty type and are set as hyperparameters. These hyperparameters allow for tuning the penalty's impact on the reward and therefore, provide fine-grained control of which aspects of the model's capabilities should be explored in-depth during the benchmarking process. Given that the tree generated in Phase~1 is used for all subsequent phases, high weights for errors and failures will decrease the overall reward at each node and therefore increase the overall difficulty level of the entire benchmarking process.
    
            In this way, Equation~\ref{eq:phase_1_reward} allows for mapping model capabilities: the more successful the model is at solving challenges, the higher the TD values for the tree's nodes, and the more MCTS is encouraged to continue exploring the search tree to find challenging areas.
    
        \subsubsection{Phase~2: Challenge Discovery}\label{subsubsec:appendix:modified_tree_search_algorithm:phase_specific_reward_functions:phase_2}
            Phase~2 shifts focus from broad capability mapping in Phase~1 to systematically identifying the model’s capability boundaries. Here, the reward function as described in Equation~\ref{eq:phase_2_reward} prioritizes the challenges where the model struggles. The reward is calculated as:
            \begin{equation*}
                R_2(s) = \lambda(1 - r_{success}) + \gamma \cdot n_{attempts} + \beta \cdot I_{fixer}
            \end{equation*}
            With $r_{success}$ being the ratio of successfully passed tests (no errors or failures). Using the complement of the success ratio assigns higher rewards to nodes where the success rate is low. Therefore, nodes that consistently expose the model’s inability to generate correct solutions receive higher rewards. The hyperparameter $\lambda$ allows for controlling how aggressively the benchmark focuses on nodes with low success rates. $n_{attempts}$ is the number of attempts it took for the model to fix a solution that had failed/errored tests. Therefore, nodes requiring multiple attempts receive higher rewards and $\gamma$ adjusts the weight given to repeated failures. Finally, $I_{fixer}$ is calculated in the same way as in Phase~1, with $\beta$ controlling the weight of the penalty for dependency on the \textit{Fixer} agent.
            
            Equation~\ref{eq:phase_2_reward} allows MCTS to explore regions of the search space where the model \textit{consistently} fails (i.e., the more the model fails at each node, the higher the node's TD value will be). As such, Phase~2 generates a refined set of nodes where the model has constantly underperformed. These nodes will be used for analysis of the underlying root causes of poor performance in Phase~3.

            
        \subsubsection{Phase~3: Comprehensive Evaluation}
            Phase~3 focuses on analyzing the root causes of model failures while using the same reward formula as in Phase~2.
            
    \subsection{Tree Expansion and Action Selection}\label{subsec:appendix:modified_tree_search_algorithm:tree_exapnsion_mechanism}
        The decision to expand a node determines how the tree grows and how we explore the search space. In Phases 1 and 2, node expansion $E(n)$ is governed by two criteria: \begin{equation}
            E(n) = \begin{cases}
                1 & \text{if } v(n) \geq \theta_p \text{ and } d(n) \leq d_{max}\\
                0 & \text{otherwise}
            \end{cases}
        \end{equation}
        where $v(n)$ is the node's TD value, $\theta_p$ is the TD value threshold of each phase, $d(n)$ is the node's depth, and $d_{max}$ is the maximum allowed depth for each phase. $\theta_p$ controls the collective difficulty of each phase, with higher thresholds indicating harder acceptance criteria for solution acceptance at each node. $d_{max}$ defines a hard limit on how deep the tree can get at each phase with higher values allowing for more in-depth analysis at each phase. When expansion is triggered, the node can be expanded in two ways: 
        \begin{equation}
        \label{eq:expnasion_mechanism}
            a_{expand}(n) = \begin{cases}
                \text{combine\_concepts}(n, n') & \text{with probability } p_e\\
                \text{increase\_difficulty}(n) & \text{with probability } 1-p_e
            \end{cases}
        \end{equation}
        where $n'$ is another selected node for concept combination and $p_e$ is the probability of which expansion action is selected and tuned based on the benchmark's desired level of exploration.
            
        As mentioned in \ref{subsubsec:appendix:modified_tree_search_algorithm:state_selection_policies:phase_3}, in Phase~3, all nodes that have a TD value higher than the set threshold $\theta_c$ are selected for analysis. Therefore in this phase, all selected nodes are expanded deterministically to create variations of challenging problems, maintaining the same concept combination and difficulty level, to thoroughly evaluate the model's behavior on low-performing nodes.

\section{Agents}\label{sec:appendix:agents}
    Even though the term ``agent" is well-established in RL literature, LLM providers maintain different interpretations of what constitutes an agent \cite{shavit2023practices, anthropic2024}. In order to have a uniform definition throughout our work, we adopt the definition from \cite{smolagents}, which characterizes LLM-based agents as ``programs where LLM outputs control the workflow.'' In this manner, our multi-agent system integrates with the search tree and MCTS through an orchestrated feedback mechanism: \textit{Problem Generators} influence expansion strategies by generating challenges while \textit{Pattern Analyzers} and \textit{Solution Evaluators} conduct phase-specific assessments that influence the reward for each node according to the phase's goal. The agents coordinate through a shared state that maintains performance history, generated problems, and evaluation results, ensuring consistency across evaluation aspects while enabling dynamic search strategy adaptation based on collected results. As mentioned in Section \ref{subsec:methodology_agents}, the model under benchmark can be configured to any of the specified agent roles, enabling fine-grained and targeted capability assessment. This flexibility makes \tool{} adaptable to diverse evaluation requirements.
    
    \subsection{Agent Responsibilities}
        \tool{} is comprised of seven agents. All of our agents utilize prompting best practices \cite{marvin2023prompt, openai_prompt_engineering, anthropic_prompt_engineering}: clearly defined roles, structured output schemas, and few-shot examples. Our agent roles are as follows:
        \begin{itemize}
            \item Challenge Designer: Generates programming challenges of specified difficulty levels targeting particular computer science concepts, following an LC-style format with clear input/output specifications and constraints.
            \item Test Generator: Generates comprehensive test suites that validate functional correctness, corner cases, and performance constraints of submitted solutions while ensuring full coverage of the challenge requirements.
            \item Problem Solver: Generates code to implement a solution for the programming challenge with an emphasis on efficiency and adherence to best practices, while handling all specified corner cases and constraints.
            \item Problem Fixer: Analyzes the outputs of program execution and implementation details (solution + tests) to fix failures for the solutions and tests.
            \item Test Validator: Evaluates generated test suites for comprehensiveness, identifying potential gaps in code coverage, missing corner cases, and opportunities for improvement.
            \item Test Error Analyzer: Performs detailed analysis of test execution failures, categorizing error patterns and providing insights into the root causes of solution failures.
            \item Solution Pattern Analyzer: Examines implemented solutions to identify algorithmic approaches, data structure usage, and implementation patterns, providing metrics for solution quality and efficiency.
        \end{itemize}
        
        \begin{figure}[h]
          \centering
          \includegraphics[width=0.9\textwidth]{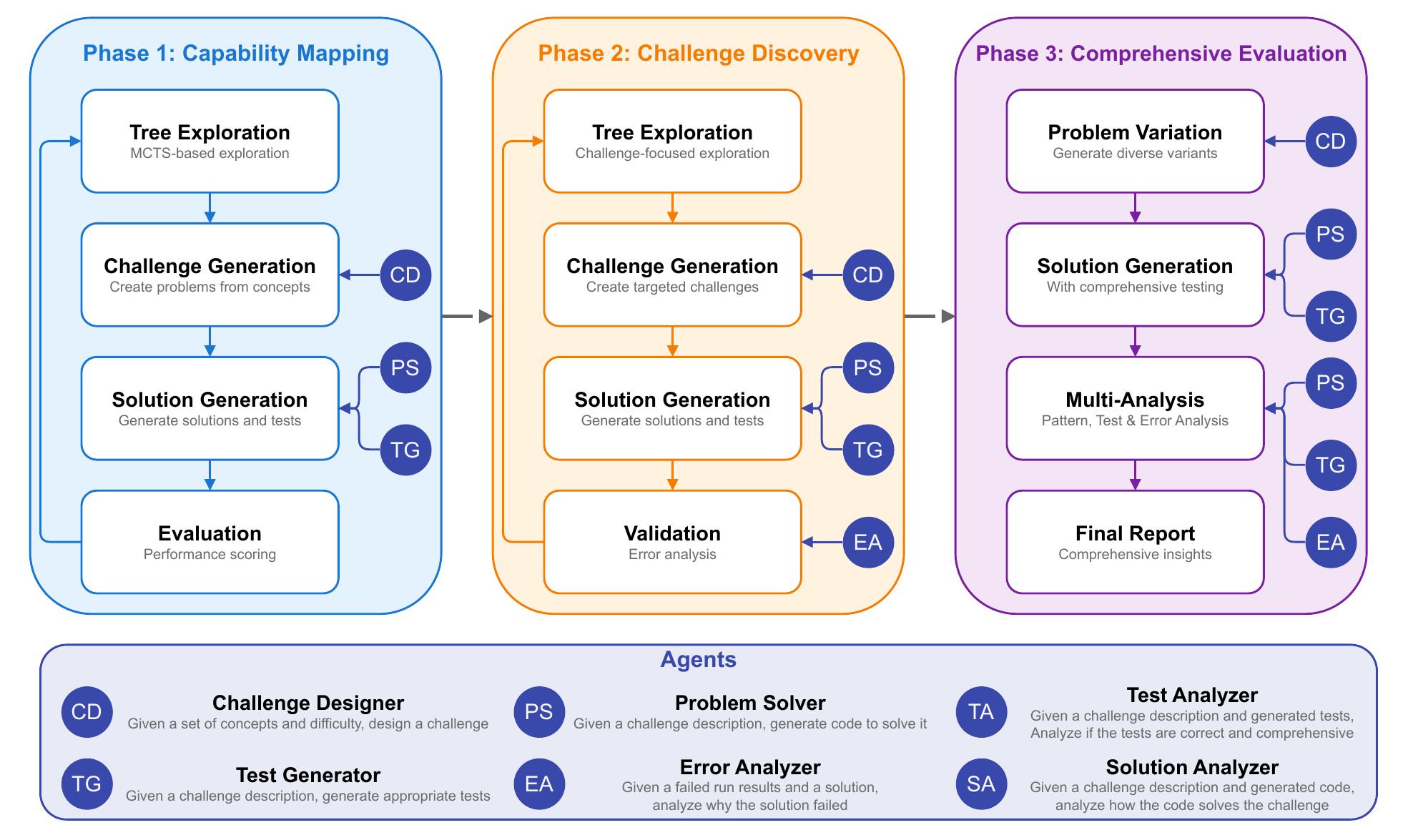}
          \caption {The agent interaction and use throughout each phase.}
          \label{fig:agent_overview}
        \end{figure}
        
        Figure~\ref{fig:agent_overview} displays a holistic view of where each agent is used throughout each phase in {\tool{}}. The complete prompt templates and configuration parameters for these agents are available in our replication package \cite{replication_package}. 
    
    \subsection{Interaction Sandbox}
\begin{figure}[h]
          \centering
          \includegraphics[width=0.9\textwidth]{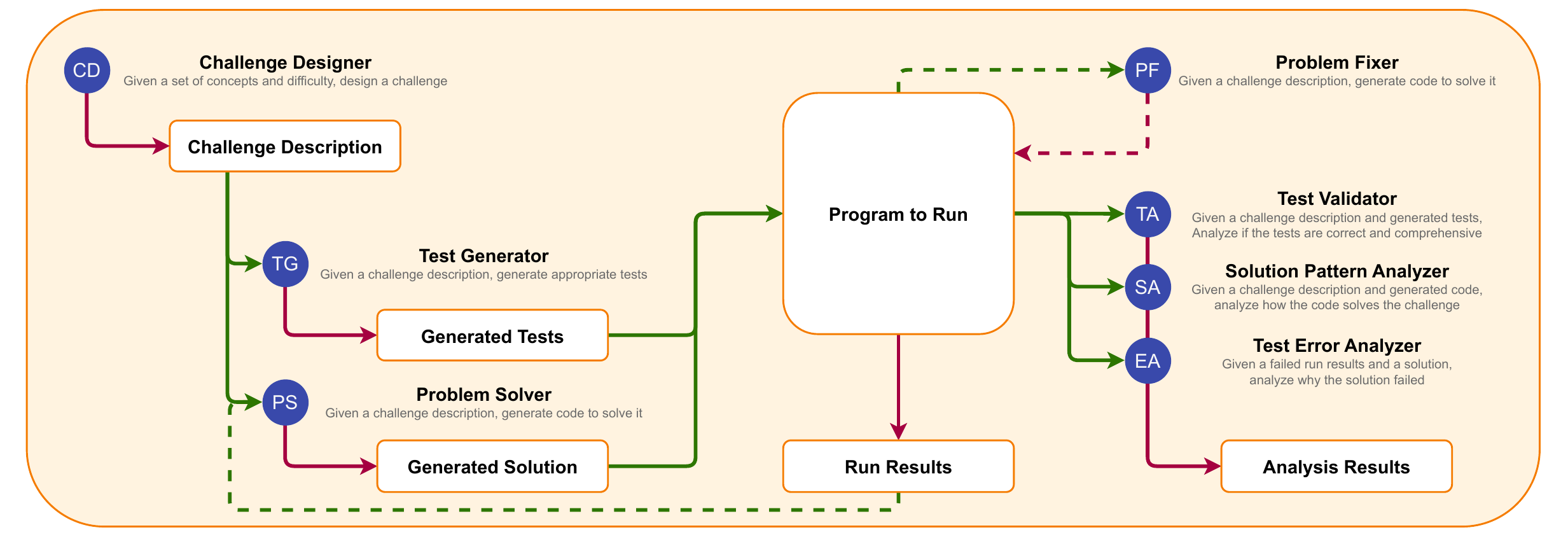}
          \caption {A diagram of the interaction between agents at each node. Green arrows indicate inputs, red arrows indicate outputs. Dashed lines indicate inputs/outputs triggered by conditions.}
          \label{fig:interaction_sandbox}
        \end{figure}

        Our aim in \tool{} is to design a general framework to evaluate LLMs on code-related tasks. Therefore, given that many LLMs lack function-calling/tool-use capabilities or may not consistently adhere to preset output formats, we designed a controlled sandbox environment that enables structured agent interactions while maintaining evaluation integrity. Figure \ref{fig:interaction_sandbox} and Algorithm \ref{alg:agent_interaction} present the sandbox architecture and the execution flow, which are instantiated for \textbf{each node} in the search tree, respectively. 
        
        The interaction cycle at each node begins with the \textit{Challenge Designer} agent, which generates LC-style programming challenges based on specified concepts and difficulty levels as shown in Listing \ref{listing:challenge_example}. These challenges serve as the primary input for two agents: \textit{Test Generator} and \textit{Problem Solver}. The Test Generator agent is tasked with generating a comprehensive test suite, while the Problem Solver agent is responsible for implementing a solution. Both agents operate independently, with access limited to the challenge description to prevent cross-contamination of their outputs. The sandbox environment then executes the generated solution against the generated test suite, capturing detailed metrics including passed tests count, failure types/counts, error types/counts, and execution traces as shown in Listing \ref{listing:run_results_example}. Upon \textbf{successful completion} of all tests, the node is marked as resolved, and control returns to the search algorithm. However, if test failures occur, \tool{} initiates an iterative feedback process.
        
        \begin{algorithm}[H] 
        \caption{Agent Interaction at Each Node}
        \label{alg:agent_interaction}
        \textbf{Input:} $C$: List of concepts, $D$: Difficulty level \\
        \textbf{Output:} $S$: Node score, $M$: Collected metrics
            \begin{algorithmic}[1]
            \renewcommand{\algorithmiccomment}[1]{\hfill #1}  
            \STATE $challenge\_description \gets \textsc{ChallengeDesigner}(C, D)$
            \STATE $g\_t \gets \textsc{TestDesigner}(challenge\_description)$ \COMMENT{Generated tests}
            \STATE $g\_s \gets \textsc{ProblemSolver}(challenge\_description)$ \COMMENT{Generated solution}
            \STATE $p\_r \gets g\_s \oplus g\_t$ \COMMENT{Combine solution and tests}
            \STATE $(Success, Run\_results) \gets \textsc{Run}(p\_r)$
            \IF{$Success$}
                \STATE $T_{validation} \gets \textsc{TestValidator}(g\_t)$ \COMMENT{Analyze the tests}
                \STATE $P_{solution} \gets \textsc{SolutionPatternAnalyzer}(g\_s)$ \COMMENT{Analyze the solution}
            \ELSE
                \FOR{$i = 1$ to $num\_attempts$}
                    \STATE $e\_f \gets \text{errors during run}$ \COMMENT{Collected errors}
                    \STATE $f\_s \gets \textsc{ProblemSolver}(g\_s, e\_f)$ \COMMENT{Generate fixed solution}
                    \STATE $E_{analysis} \gets \textsc{TestErrorAnalyzer}(g\_s, g\_t, e\_f)$ \COMMENT{Analyze the errors}
                    \STATE $p\_r \gets f\_s \oplus g\_t$
                    \STATE $(Success, Run\_results) \gets \textsc{Run}(p\_r)$
                    \IF{$Success$}
                        \STATE \textbf{break}
                    \ENDIF
                \ENDFOR
                \IF{not $Success$}
                    \STATE $fixed\_p\_r \gets \textsc{ProblemFixer}(p\_r)$ \COMMENT{Use \textit{Problem Fixer} agent}
                    \STATE $(Success, Run\_results) \gets \textsc{Run}(fixed\_p\_r)$
                \ENDIF
            \ENDIF
            \STATE $M \gets (T_{validation}, P_{solution}, E_{analysis}, Run\_results)$ \COMMENT{Store all run results}
            \STATE \textbf{return} $S, M$
            \end{algorithmic}
        \end{algorithm}
                
        In the feedback phase, the Problem Solver agent receives execution results and error details, attempting to correct the solution. This approach serves two purposes: it accounts for the stochasticity in LLM performance as described in \ref{subsec:appendix:modified_tree_search_algorithm:state_selection_policies} (if the model fails, it is provided with context to revise its response) while also evaluating the model's capability to learn from feedback. If multiple solution attempts fail to resolve the issues within a predetermined limit, \tool{} calls the \textit{Problem Fixer} agent. The Problem Fixer, which may be either the model under benchmark itself or a separate model depending on evaluation requirements, receives comprehensive context 
        including the challenge description, implementation history, test suite, and failure results. Allowing the model to have access to all of the previously collected information, enables assessment of program repair capabilities of the model by providing it with full contextual information. The cycle of testing and refinement continues until success is achieved (the resulting \textit{program to run}, $p\_r$, executes with no errors and failures) within a predetermined limit. Otherwise, the node is marked as failed.
        
        \begin{lstlisting}[caption={Challenge Designer output for a set of concepts and difficulty level}, label={listing:challenge_example}]
title: "Even or Odd"
concepts:
    - "conditionals"
    - "functions"
difficulty: "very easy"
description: "## Even or Odd
  Write a function that takes an integer as input and determines whether the number is even or odd. The function should return the string \"Even\" if the number is even, and \"Odd\" if the number is odd.

  ### Input:
  - n: An integer (-10^9 <= n <= 10^9)

  ### Output:
  - A string \"Even\" or \"Odd\" based on the parity of the input integer.

  ### Constraints:
  - -10^9 <= n <= 10^9

  ### Examples:
  1. Input: n = 4  
     Output: \"Even\"  
     Explanation: The number 4 is divisible by 2, hence it is even.

  2. Input: n = 7  
     Output: \"Odd\"  
     Explanation: The number 7 is not divisible by 2, hence it is odd.

  ### Relevance to Conditionals and Functions:
  This problem tests the understanding of basic conditionals, as the solution requires checking the remainder when the number is divided by 2. It also reinforces the use of functions for encapsulating logic, demonstrating how to structure a simple program."
        \end{lstlisting}

        \begin{lstlisting}[caption={an example of the run results for a node},label={listing:run_results_example}]
problem_statement: "## Even or Odd..."
success= True
tests_passed: 10
tests_failed: 2
tests_errored: 0
fixed_by_problem_fixer=false
data_trail:
    {
        attempt_1:
            {
                test_cases: "import unittest\n\n...",
                solution_code: "def solution(...)",
                output: "'Tests failed. Output:\n\n....F.....\n=..",
            },
        attempt_2:
            {
                test_cases: "import unittest\n\n...",
                solution_code: "def solution(...)",
                output: "All Tests passed",   
            },
    }
        \end{lstlisting}

        Phase~3 also introduces additional diagnostic analyses through the analyzer agents. These agents process the complete execution history of each node, enabling detailed analysis of failure modes and performance patterns. Once Phase~3 is finished, all the results are used to create comprehensive reports about the model's capabilities. 
        

\section{Metrics}\label{sec:appendix:metrics}
    \tool{} employs multiple sets of metrics to thoroughly capture and analyze the model's code generation capabilities and limitations. In this part, we will detail our defined metrics.

    \subsection{Tree Exploration and Traversal}
        The first set of metrics helps us understand how the model explores and navigates the tree throughout the entire evaluation process.
        
        \textbf{How does the model traverse the tree?}\\
        We track the distribution and connectivity of explored concepts through structural metrics:
        $$N(c) = \sum_{n \in \text{nodes}} 1_{[c \in \text{concepts}(n)]}$$
        $$N(d) = \sum_{n \in \text{nodes}} 1_{[d \in \text{difficulties}(n)]}$$
        with $N(c)$ and $N(d)$ being the number of times each concept and each difficulty was encountered throughout the entire tree. The node distribution across concepts and difficulties provide a broad view of where the model succeeds and struggles: the greater the number of nodes associated with each concept and difficulty level, the less successful the model has been in addressing related challenges. Consequently, additional nodes were generated to better identify and isolate the problematic areas. 
        
        This is complemented by the branching factor at each node:
        $$B(n) = \frac{\text{children}(n)}{|N|}$$
        where $\text{children}(n)$ is the number of children of node $n$ and $N$ represents the total number of nodes. Nodes with higher branching factors have more children compared to the other nodes and therefore have been more challenging for the model.

        The convergence rate $C(n)$ measures the stability of a model’s performance at each node $n$ by measuring the difference between consecutive TD values:
        $$C(n) = |v_{t+k} - v_t| < \epsilon \text{ for } k=1,\ldots,K$$
        where $v_t$ represents the node's TD value at attempt $t$. The convergence rate reflects how drastically the model’s output changes between attempts. A phase is terminated when all nodes exhibit convergence rates below a predefined threshold $\epsilon$ for $K$ consecutive attempts. A lower convergence rate indicates greater stability, meaning the model’s performance has plateaued at node $n$. When this condition holds across all nodes in a phase, the phase is deemed to have been sufficiently explored and the next phase begins.

    \subsection{Identifying Model Capabilities}
        These metrics assess the model's performance across different concepts and difficulty levels.
        
        \textbf{What concepts does the model understand well?}\\
        The primary measure of concept mastery is the success rate:
        $$SR(c) = \frac{1}{N(c)} \sum_{n \in N} \text{success}(n)$$
        where $\text{success}(n)$ is 1 if the model has successfully passed the challenge at node $n$ and 0 otherwise as shown in Listing~\ref{listing:run_results_example}. 
        
        Similarly, we measure the model's success rate at each difficulty level:
        $$SR(d) = \frac{1}{N(d)} \sum_{n \in N} \text{success}(n)$$
        
        To understand the effort required for solving a challenge related to a concept $c$, we measure the average number of attempts regardless of success or failure:
        $$A(c) = \frac{1}{N(c)} \sum_{n \in N(c)} \text{Attempts}(n)$$
        with $\text{Attempts}(n)$ representing the number of attempts made at node $n$ as shown in Listing~\ref{listing:run_results_example}.
        
        These three metrics alongside each other, indicate how well the model performs in solving challenges for each specific concept/difficulty with the average number of attempts indicating how many times the model encountered errors while solving the challenge. High success rates and low number of attempts indicate a high capability (the challenge was solved with a low number of errors and attempts) while lower success rates and higher number of attempts indicate struggles in solving challenges with that specific concept/difficulty.

        
        \textbf{When was the model unable to solve the challenge?}\\
        The fixer intervention rate indicates when the model requires external help:
        $$F(c) = \frac{1}{N(c)} \sum_{n \in N(c)} 1_{[I_{fixer}(n)]}$$
        with $I_{fixer}$ being 1 if the \textit{Problem Fixer} agent was used at each node $n$ and 0 otherwise.

        \textbf{How well does the model perform at program repair?}\\
        As shown in Algorithm~\ref{alg:agent_interaction} the \textit{Problem Fixer} agent is only used when the model fails in all of its attempts to solve the challenge. This is caused by either incorrect solutions or incorrect tests. Therefore, we can measure the model's program repair capabilities for each concept by tracking whether the use of \textit{Problem Fixer} resulted in success:
        $$R(c) = \frac{\sum_{n \in N(c)} 1_{[\text{success}(n)]}}{\sum_{n \in N(c)} 1_{[I_{\text{fixer}}(n)]}}$$
        where $\text{success}(n)$ is 1 if the model has successfully passed the challenge at node $n$ and 0 otherwise after the \textit{Problem Fixer} intervention at node $n$.

    \subsection{Understanding Model Behavior}
        These diagnostic metrics help identify and characterize the model's behavior (how it solves the challenges and how it fails).
        
        \textbf{What types of solutions does the model prefer?}\\
        The distribution of solution patterns across concepts shows how many times the model has used a specific solution for each concept $c$:
        $$SP(p,c) = \frac{\text{count}(p,c)}{N(P)}$$
        with $N(P)$  being the number of identified patterns throughout the entire tree and patterns indicating algorithmic approaches, data structure usage, and implementation patterns.
        
        \textbf{What solution patterns correlate with success?}\\
        Pattern effectiveness quantifies which solutions the model executes successfully, helping identify its preferred problem-solving strategies:
        $$PE(p) = \frac{\sum_{n \in N(p)} 1_{[\text{success}(n)]}}{N(P)}$$
        Conversely, using failure rate $(1 - success(n))$ instead of success rate quantifies the patterns the model struggles with the most.

        \textbf{How does the model generate tests?}\\
        Test validation scores test suite quality:
        $$TV(v,c) = \frac{\text{count}(v,c)}{N(V)}$$
        where $N(V)$ is the number of identified validation issues throughout the entire tree with $v$ being an identified validation issue with validation issues including analyses on missing, incorrect, coverage, and corner case issues for each generated test suite for each concept $c$.
        
        \textbf{What are the common failure patterns?}\\
        Error pattern distribution by concept shows where exactly the model has failed in solving the challenges related to that concept $c$:
        $$EP(e,c) = \frac{\text{count}(e,c)}{N(E)}$$
        where $N(E)$ is the number of identified errors throughout the entire tree with $e$ being an error that was raised during the execution of the program.

    This set of metrics enables us to understand not just what the model can do, but how it performs and where it struggles.

\section{Detailed Analysis}\label{sec:appendix:results}
    \subsection{Concepts}\label{subsec:appendix:results:concepts}
        As mentioned earlier, this study benchmarks large language models to evaluate their proficiency in understanding and implementing fundamental computer science concepts. Below, we provide a concise explanation of each concept and what we expect the models to achieve in tasks involving these concepts.
        \begin{itemize}
            \item \textbf{Loops}: A loop is a control structure that repeatedly executes a block of code as long as a specified condition is true. Examples include \texttt{for}, \texttt{while}, and \texttt{do-while} loops. As such the models should:
            \begin{itemize}
                \item Correctly implement loops to traverse data structures or repeat operations.
                \item Optimize loop usage for efficiency and avoid common pitfalls such as infinite loops.
            \end{itemize}
            \item \textbf{Conditionals}: Conditionals are control structures that execute specific code blocks based on boolean conditions. Examples include \texttt{if}, \texttt{else}, and \texttt{else if} statements. We expect the model to:
            \begin{itemize}
                \item Accurately implement conditionals to manage decision-making logic.
                \item Handle edge cases and ensure logical correctness when combining multiple conditions.
            \end{itemize}
            \item \textbf{Functions}: Functions are reusable blocks of code that perform a specific task, defined by a name, parameters, and a return value. The models should:
            \begin{itemize}
                \item Design modular and reusable functions.
                \item Handle parameter passing and scope effectively.
            \end{itemize}
            \item \textbf{Data Structures}: Data structures organize and store data to facilitate efficient access and modification. Examples include arrays, linked lists, stacks, queues, and trees. The models should:
            \begin{itemize}
                \item Choose appropriate data structures for given problems.
                \item Implement and manipulate data structures accurately and handle edge cases.
            \end{itemize}
            \item \textbf{Algorithms (logic)}: Step-by-step procedures for solving problems or performing computations. As such, the models should:
            \begin{itemize}
                \item Devise efficient algorithms to address specified problems.
                \item Optimize time and space complexity, demonstrating an understanding of computational trade-offs.
            \end{itemize}
            \item \textbf{Error Handling}: Error handling involves detecting, managing, and responding to runtime errors. As such, the models should:
            \begin{itemize}
                \item Implement robust error-handling mechanisms, including exception handling and validation.
            \end{itemize}
            \item \textbf{Recursion}: Recursion is a technique where a function calls itself to solve a problem by breaking it into smaller sub-problems. As such, the models should:
            \begin{itemize}
                \item Correctly implement recursive functions, ensuring termination through base cases.
                \item Optimize recursion to avoid excessive memory usage and stack overflow issues.
            \end{itemize}
            \item \textbf{Sorting}: Sorting involves arranging data in a specific order, such as ascending or descending such as quicksort, mergesort, and bubble sort. As such, the models should:
            \begin{itemize}
                \item Implement sorting algorithms correctly and select appropriate algorithms for the given data size and constraints.
            \end{itemize}
            \item \textbf{Searching}: Searching involves finding specific elements in a dataset such as linear search, binary search, and hash-based lookups. As such, the models should:
            \begin{itemize}
                \item Apply efficient search techniques suited to the dataset’s structure.
                \item Ensure correctness and handle cases where the element is not present.
            \end{itemize}
            \item \textbf{Dynamic Programming}: Dynamic programming is a technique for solving complex problems by breaking them into overlapping sub-problems and solving each sub-problem only once. We expect the models to:
            \begin{itemize}
                \item Develop dynamic programming solutions to problems requiring optimization.
                \item Demonstrate the ability to use memoization or tabulation correctly.
            \end{itemize}
        \end{itemize}
        
        These concepts are foundational to CS and cover the essential problem-solving skills required to implement solutions and tests for a problem. By benchmarking models on these concepts, we aim to assess their ability to generalize to unseen tasks based on single concepts and concept combinations critical for coding and reasoning. The concepts for benchmarking are modifiable, meaning that they can be changed to any desired topic, allowing {\tool{}} to be used in more specific scenarios and subjects (e.g., instead of foundational concepts, implementation patterns and challenges closer to LC challenges such as ``Two Sum'', ``Valid Sudoku'', etc. can be chosen). 

    \subsection{Combination of Concepts}\label{subsec:appendix:results:combinations_of_concepts}
        In real-world programming scenarios, the implementation of solutions rarely require implementing isolated, single concepts. Instead, they require the integration of multiple concepts to address complex problems effectively. For example, developing a functional application often involves combining loops for iteration, conditionals for decision making, and data structures to organize information. In addition, advanced challenges frequently require recursion, algorithms for processing logic, and error handling to ensure that the program does not fail when it encounters unexpected inputs or conditions.

        Therefore, to simulate real-world programming scenarios, {\tool{}} generates challenges that combine these core concepts into unified problems. This allows us to evaluate a model's capabilities to 
        synthesize knowledge across programming concepts. For example, a single problem might require using dynamic programming alongside data structures for optimal solutions or using sorting and searching techniques to manage/query datasets. This approach ensures that the model can demonstrate competency in scenarios requiring cross-concept integration. As such, failure to solve problems involving multiple concepts is an indicator of deficiencies in one or more of the constituent concepts. Such failures signal areas where the model struggles to integrate distinct methodologies or lacks a deep understanding of specific concepts. For instance, if a model fails a task combining functions and error handling, it might reflect difficulties in managing exceptions within modularized code. In this manner, {\tool{}} can investigate these failures further by identifying the exact concepts or combinations responsible for failures.

        Alongside combining concepts, we also use a range of difficulty levels: very easy, easy, medium, hard, and very hard in order to perform fine-grained analysis of the model's capabilities. This enables us to assess performance not only on single concepts and their combinations but also on different complexities of these problems. For example, a model might perform well on easier problems related to a concept or group of concepts but fail on medium or hard ones, revealing limitations in its ability to scale solutions to more challenging scenarios. 
        
        By probing models across a variety of concept combinations and difficulty levels, we gain a comprehensive understanding of their strengths and weaknesses and gain valuable insights into their overall code generation capabilities by pinpointing root causes and systematically evaluating a model's limitations.
    
    \subsection{Detailed Analysis of Results}\label{subsec:appendix:analysis:detailed_results}
        Table \ref{tab:detailed_difficulty_performance} and \ref{tab:detailed_concept_performance} show the average success rate and average intervention rate for each of the models under study, across concepts and difficulties, respectively. The metrics presented here, are averaged from the values throughout the entire tree at the end of the benchmarking process, for 3 independent benchmarking runs for each model, and are not phase-specific.
        \begin{table}[h]
        \centering
        \caption{Model Performance by Difficulty. Colors indicate performance: green (good) to red (poor). Higher values for intervention rates indicate more usage of the \textit{Fixer} agent.}
        \label{tab:detailed_difficulty_performance}
        \resizebox{\textwidth}{!}{%
            \begin{tabular}{lc|c c c|c c c|c c c|c c c|c}
            \toprule
            \textbf{Difficulty} & \multicolumn{2}{c}{4o} & & \multicolumn{2}{c}{4o-M} & & \multicolumn{2}{c}{L405} & & \multicolumn{2}{c}{L70} & & \multicolumn{2}{c}{L8} \\
             & Avg Succ. Rate & Avg Inter. & & Avg Succ. Rate & Avg Inter. & & Avg Succ. Rate & Avg Inter. & & Avg Succ. Rate & Avg Inter. & & Avg Succ. Rate & Avg Inter.\\
            \midrule
            Very easy & \cellcolor{goodgreen!25}0.83 & \cellcolor{badred!66}4.67 & & \cellcolor{goodgreen!25}0.83 & \cellcolor{midyellow!25}1.00 & & \cellcolor{goodgreen!25}0.85 & \cellcolor{badred!56}3.67 & & \cellcolor{midyellow!66}0.42 & \cellcolor{badred!56}3.00 & & \cellcolor{badred!62}0.19 & \cellcolor{midyellow!25}1.67\\
            Easy & \cellcolor{goodgreen!25}0.73 & \cellcolor{badred!56}2.00 & & \cellcolor{midyellow!61}0.63 & \cellcolor{midyellow!25}1.00 & & \cellcolor{goodgreen!25}0.72 & \cellcolor{badred!56}2.33 & & \cellcolor{badred!66}0.22 & \cellcolor{badred!56}2.33 & & \cellcolor{badred!62}0.22 & \cellcolor{midyellow!25}1.00\\
            Medium & \cellcolor{midyellow!61}0.44 & \cellcolor{midyellow!25}1.00 & & \cellcolor{midyellow!61}0.35 & \cellcolor{midyellow!25}1.00 & & \cellcolor{midyellow!61}0.43 & \cellcolor{midyellow!56}1.33 & & \cellcolor{badred!66}0.16 & \cellcolor{badred!56}2.00 & & \cellcolor{badred!90}0.03 & \cellcolor{midyellow!25}0.00\\
            Hard & \cellcolor{midyellow!61}0.33 & \cellcolor{badred!56}2.00 & & \cellcolor{badred!62}0.21 & \cellcolor{midyellow!56}1.50 & & \cellcolor{midyellow!61}0.50 & \cellcolor{midyellow!25}1.00 & & \cellcolor{badred!90}0.00 & \cellcolor{midyellow!25}0.00 & & \cellcolor{badred!90}0.00 & \cellcolor{midyellow!25}0.00\\
            Very hard & \cellcolor{badred!62}0.26 & \cellcolor{badred!56}2.00 & & \cellcolor{badred!62}0.24 & \cellcolor{midyellow!25}1.00 & & \cellcolor{badred!62}0.30 & \cellcolor{midyellow!25}0.00 & & \cellcolor{badred!90}0.00 & \cellcolor{midyellow!25}0.00 & & \cellcolor{badred!90}0.00 & \cellcolor{midyellow!25}0.00\\
            \bottomrule
            \end{tabular}}
        \end{table}

        Looking at the performance data across all models, we observe a clear hierarchy in both success rates and number of interventions. Starting with the model performance by difficulty level, there's a consistent degradation in success rates as difficulty increases across all models. As expected, the ``very easy'' difficulty level shows the highest success rates for all models. The success rates steadily decline to much lower values at ``very hard'' difficulties. The success rates of {\lseventyb} and {\leightb} even on the ``very easy'' difficulty level compared to the other models, already indicate the limited capability of these models given the number of their parameters which we discuss in depth in \ref{subsec:appendix:detailed_analysis:effects_of_scale}.

        \begin{table}[h]
            \centering
            \caption{Model Performance by Concept. Colors indicate performance: green (good) to red (poor). Higher values for intervention rates indicate more usage of the \textit{Fixer} agent.}
            \label{tab:detailed_concept_performance}
            \resizebox{\textwidth}{!}{%
                \begin{tabular}{lc|c c c|c c c|c c c|c c c|c}
                \toprule
                \textbf{Concept} & \multicolumn{2}{c}{4o} & & \multicolumn{2}{c}{4o-M} & & \multicolumn{2}{c}{L405} & & \multicolumn{2}{c}{L70} & & \multicolumn{2}{c}{L8} \\
                 & Avg Succ. Rate & Avg Inter. & & Avg Succ. Rate & Avg Inter. & & Avg Succ. Rate & Avg Inter. & & Avg Succ. Rate & Avg Inter. & & Avg Succ. Rate & Avg Inter.\\
                \midrule
                Loops & \cellcolor{goodgreen!25}0.53 & \cellcolor{badred!56}2.50 & & \cellcolor{midyellow!61}0.48 & \cellcolor{badred!56}2.00 & & \cellcolor{goodgreen!25}0.55 & \cellcolor{badred!56}3.00 & & \cellcolor{badred!66}0.21 & \cellcolor{midyellow!25}1.00 & & \cellcolor{badred!90}0.10 & \cellcolor{midyellow!25}1.00\\
                Conditionals & \cellcolor{midyellow!61}0.46 & \cellcolor{badred!66}4.33 & & \cellcolor{midyellow!61}0.45 & \cellcolor{badred!56}2.00 & & \cellcolor{midyellow!61}0.48 & \cellcolor{badred!56}2.33 & & \cellcolor{badred!66}0.32 & \cellcolor{badred!56}3.00 & & \cellcolor{badred!90}0.18 & \cellcolor{midyellow!25}2.00\\
                Functions & \cellcolor{goodgreen!25}0.60 & \cellcolor{badred!56}2.67 & & \cellcolor{midyellow!61}0.42 & \cellcolor{midyellow!25}1.50 & & \cellcolor{goodgreen!25}0.56 & \cellcolor{badred!56}2.00 & & \cellcolor{badred!66}0.24 & \cellcolor{midyellow!25}2.00 & & \cellcolor{badred!66}0.23 & \cellcolor{midyellow!25}2.00\\
                Data Struct. & \cellcolor{midyellow!61}0.45 & \cellcolor{badred!66}4.33 & & \cellcolor{midyellow!61}0.44 & \cellcolor{midyellow!25}1.00 & & \cellcolor{midyellow!61}0.51 & \cellcolor{badred!66}3.33 & & \cellcolor{badred!66}0.24 & \cellcolor{midyellow!25}1.00 & & \cellcolor{badred!90}0.11 & \cellcolor{midyellow!25}0.00\\
                Algorithms & \cellcolor{midyellow!61}0.49 & \cellcolor{badred!66}4.50 & & \cellcolor{midyellow!61}0.48 & \cellcolor{midyellow!25}1.00 & & \cellcolor{goodgreen!25}0.59 & \cellcolor{badred!56}3.00 & & \cellcolor{badred!66}0.33 & \cellcolor{midyellow!25}1.00 & & \cellcolor{badred!90}0.17 & \cellcolor{midyellow!25}1.00\\
                Error Hand. & \cellcolor{midyellow!61}0.47 & \cellcolor{midyellow!25}1.67 & & \cellcolor{midyellow!61}0.49 & \cellcolor{midyellow!25}1.00 & & \cellcolor{goodgreen!25}0.61 & \cellcolor{badred!56}2.33 & & \cellcolor{badred!66}0.27 & \cellcolor{midyellow!25}1.50 & & \cellcolor{badred!90}0.13 & \cellcolor{midyellow!25}1.00\\
                Recursion & \cellcolor{goodgreen!25}0.52 & \cellcolor{badred!56}2.00 & & \cellcolor{midyellow!61}0.43 & \cellcolor{midyellow!25}1.00 & & \cellcolor{goodgreen!25}0.57 & \cellcolor{badred!66}3.33 & & \cellcolor{badred!66}0.29 & \cellcolor{midyellow!25}1.00 & & \cellcolor{badred!90}0.20 & \cellcolor{midyellow!25}1.00\\
                Sorting & \cellcolor{goodgreen!25}0.52 & \cellcolor{midyellow!25}1.00 & & \cellcolor{badred!66}0.39 & \cellcolor{midyellow!25}0.00 & & \cellcolor{goodgreen!25}0.55 & \cellcolor{badred!56}2.33 & & \cellcolor{badred!66}0.31 & \cellcolor{midyellow!25}1.00 & & \cellcolor{badred!90}0.13 & \cellcolor{midyellow!25}1.00\\
                Searching & \cellcolor{goodgreen!25}0.58 & \cellcolor{midyellow!25}1.50 & & \cellcolor{midyellow!61}0.47 & \cellcolor{midyellow!25}1.50 & & \cellcolor{goodgreen!25}0.60 & \cellcolor{badred!66}3.33 & & \cellcolor{badred!66}0.30 & \cellcolor{midyellow!25}1.00 & & \cellcolor{badred!90}0.20 & \cellcolor{midyellow!25}1.00\\
                Dyn. Prog. & \cellcolor{badred!66}0.32 & \cellcolor{badred!56}2.50 & & \cellcolor{badred!66}0.34 & \cellcolor{midyellow!25}0.00 & & \cellcolor{midyellow!61}0.49 & \cellcolor{badred!66}3.33 & & \cellcolor{badred!66}0.23 & \cellcolor{midyellow!25}1.00 & & \cellcolor{badred!90}0.08 & \cellcolor{midyellow!25}1.00\\
                \bottomrule
                \end{tabular}}
        \end{table}

        In terms of concept mastery, we see varying performance across models. {\gfouro} performs best on ``functions" and ``searching" challenges, while struggling with ``dynamic programming". {\gfouromini} shows more consistent performance across concepts but with lower overall success rates. {\lfourhundredfiveb} demonstrates solid capabilities on ``error handling" and ``searching" challenges while also struggling with ``conditionals", and similar to {\gfouro} and {\gfouromini}, on ``dynamic programming". The smaller Llama models ({\lseventyb} and {\leightb}) show significantly lower success rates across all concepts, with {\leightb} particularly struggling with success rates mostly below 0.20.
        
        \begin{figure}[h]
            \centering
                \subfigure[Node distribution for {\gfouro}]{%
                    \includegraphics[width=\textwidth]{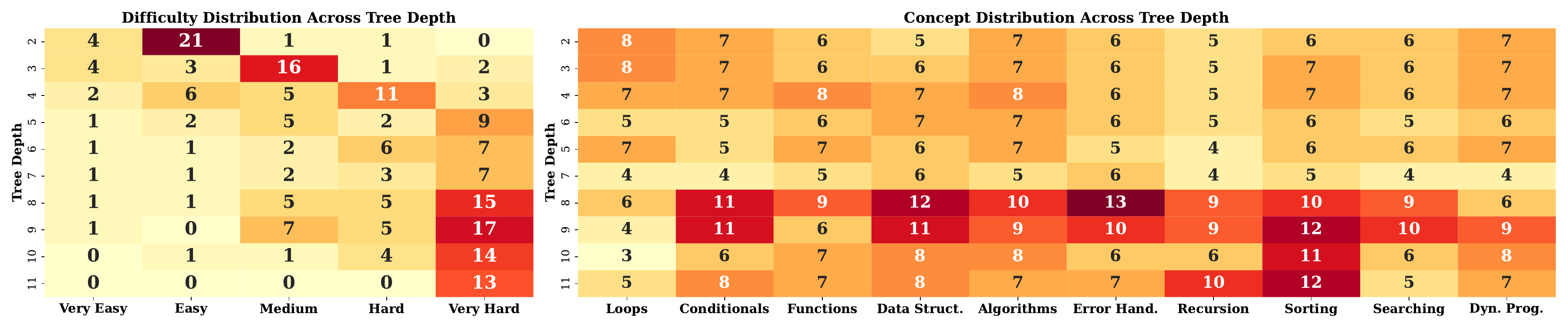}
                    \label{fig:4o_tree_metrics}
                }\\
                \subfigure[Node distribution for {\lfourhundredfiveb}]{%
                    \includegraphics[width=\textwidth]{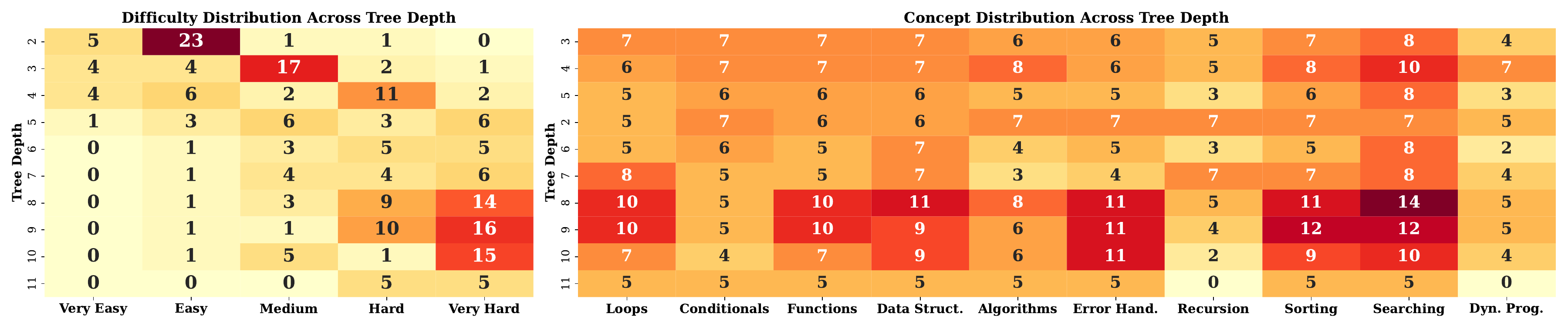}
                    \label{fig:l405_tree_metrics}
                }
            \caption{Node distributions for 4o and L-405b averaged over 3 independent runs. The numbers in each cell indicate the number of nodes.}
            \label{fig:combined_tree_metrics}
        \end{figure}
        Both {\gfouro} and {\lfourhundredfiveb} stand out with notably higher intervention rates compared to other models, especially at the ``very easy'' difficulty level (4.67 and 3.67, respectively). This is particularly interesting given that these models also maintain the highest success rates. Investigating node distributions helps explain these patterns with Figure \ref{fig:4o_tree_metrics} and \ref{fig:l405_tree_metrics} displaying the distribution of nodes in each depth per concept and difficulty for {\gfouro} and {\lfourhundredfiveb}, respectively. Both models quickly progress beyond ``very easy'' difficulty challenges, as evidenced by their node distributions (15 and 14 nodes at ``very easy" for {\gfouro} and {\lfourhundredfiveb}, respectively). As such, the high number of interventions at lower difficulties are due to smaller sample sizes at these levels combined with specific challenging cases requiring multiple interventions. On the other hand, we can observe that both {\gfouro} and {\lfourhundredfiveb} have high intervention rates for challenges related to ``conditionals", ``data structures", ``algorithms", and ``dynamic programming". Looking at the distributions of nodes per concept as shown in Figure \ref{fig:4o_tree_metrics} and \ref{fig:l405_tree_metrics} reveals that these concepts also have a high number of nodes in the deeper parts of the tree, meaning that \tool{} has identified that these concepts at high complexities have shown to be challenging for the models and has focused on these areas in order to thoroughly analyze models' capabilities.
        
        Since interventions in \tool{} are performed by the model itself through the \textit{Problem Fixer}, the combination of success rate and number of interventions effectively measure the model's program repair capabilities. {\gfouro} and {\lfourhundredfiveb} demonstrate strong program repair abilities with both high intervention and success rates, for example, at ``very easy" difficulty, {\gfouro} shows 4.67 interventions with 0.83 success rate, {\lfourhundredfiveb} shows 3.67 interventions with 0.85 success rate. As such, we can observe that when these models encounter failures, they can effectively analyze their own code, understand test failures, and implement successful fixes. This program repair capability persists even at higher difficulty levels, though with decreasing effectiveness. On the other hand, {\lseventyb} and {\leightb} have a lower number of interventions but significantly lower success rates as well. Furthermore, their success rates remain low despite interventions. For example, {\leightb} shows minimal interventions across ``very easy'' and ``easy'' but maintains very low success rates (0.19 for ``very easy", dropping to 0.00 after ``easy"). This indicates that even when given full context - including the original solution, test cases, and error outputs - these models struggle to identify and fix problems in their generated code.
        
        \begin{figure}[h]
            \centering
            \subfigure[Error pattern distribution for {\gfouro}]{
                \includegraphics[width=0.45\textwidth]{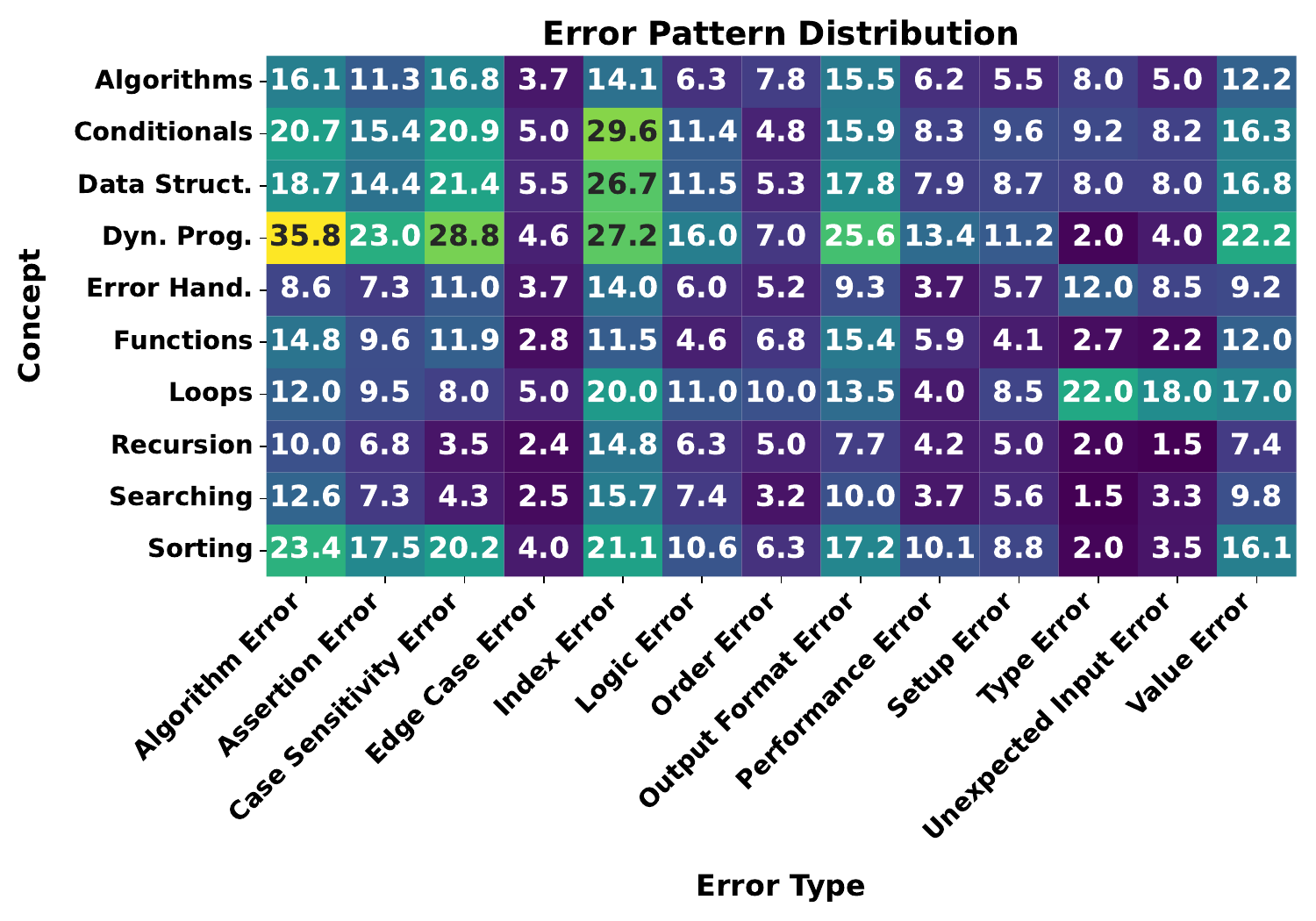}
                \label{fig:4o_error_pattern}
            }
            \hfill
            \subfigure[Error pattern distribution for {\lfourhundredfiveb}]{
                \includegraphics[width=0.45\textwidth]{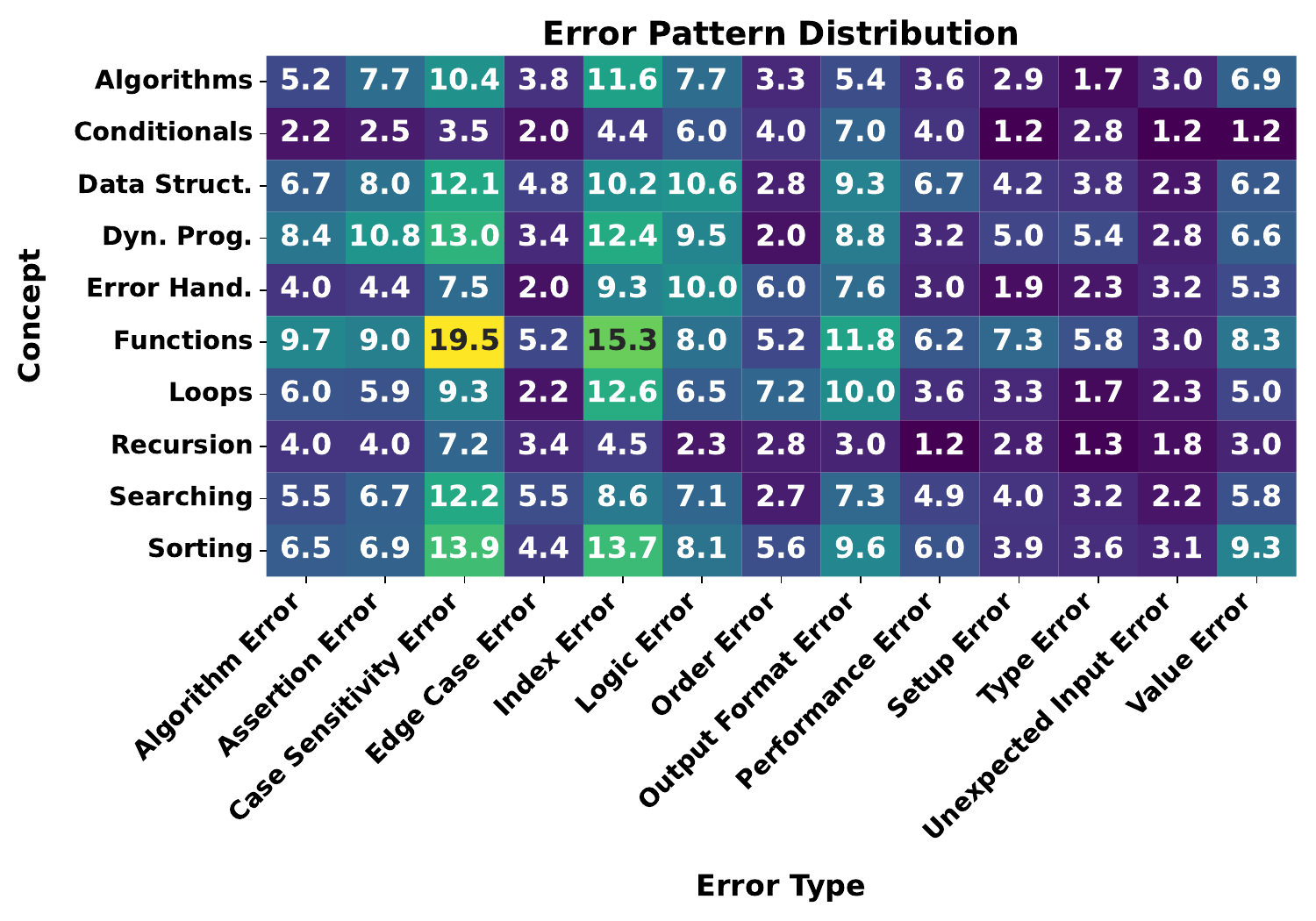}
                \label{fig:l405_error_pattern}
            }
            \caption{Error pattern distributions for 4o and L-405b averaged over 3 runs. The numbers in each cell indicate the number of times each error was raised.}
            \label{fig:4o_l405_error_pattern}
        \end{figure}
        Figure \ref{fig:4o_l405_error_pattern} shows the error pattern analysis per concept for {\gfouro} and {\lfourhundredfiveb}. The numbers in each cell represent the average occurrence of each error type per concept across 3 independent runs. {\gfouro} shows significantly higher errors in algorithm implementations, particularly in challenges related to ``dynamic programming" where algorithm implementation errors peak at 35.8 occurrences and errors related to case sensitivity peak at 28.8. Index error rates in generated code for challenges involving ``conditional" and ``data structure" concepts (29.6 and 26.7 occurrences respectively) further demonstrate {\gfouro}'s specific struggle with complex pointer and array manipulations. {\lfourhundredfiveb}'s errors, on the other hand, are mainly in ``function" implementation challenges (19.5 occurrences for case sensitivity errors, 15.3 for index errors). We can observe that {\lfourhundredfiveb} maintains consistent performance across most tested concepts, with notably lower error rates in recursive implementation challenges (consistently below 4.0 occurrences) compared to {\gfouro}. However, similar to {\gfouro}, {\lfourhundredfiveb} also struggles with ``dynamic programming", ``sorting", and ``data structure" challenges.
        
\begin{figure}[!htb]
            \centering
            \subfigure[Test validation issues distribution for {\gfouro}]{
                \includegraphics[width=0.45\textwidth]{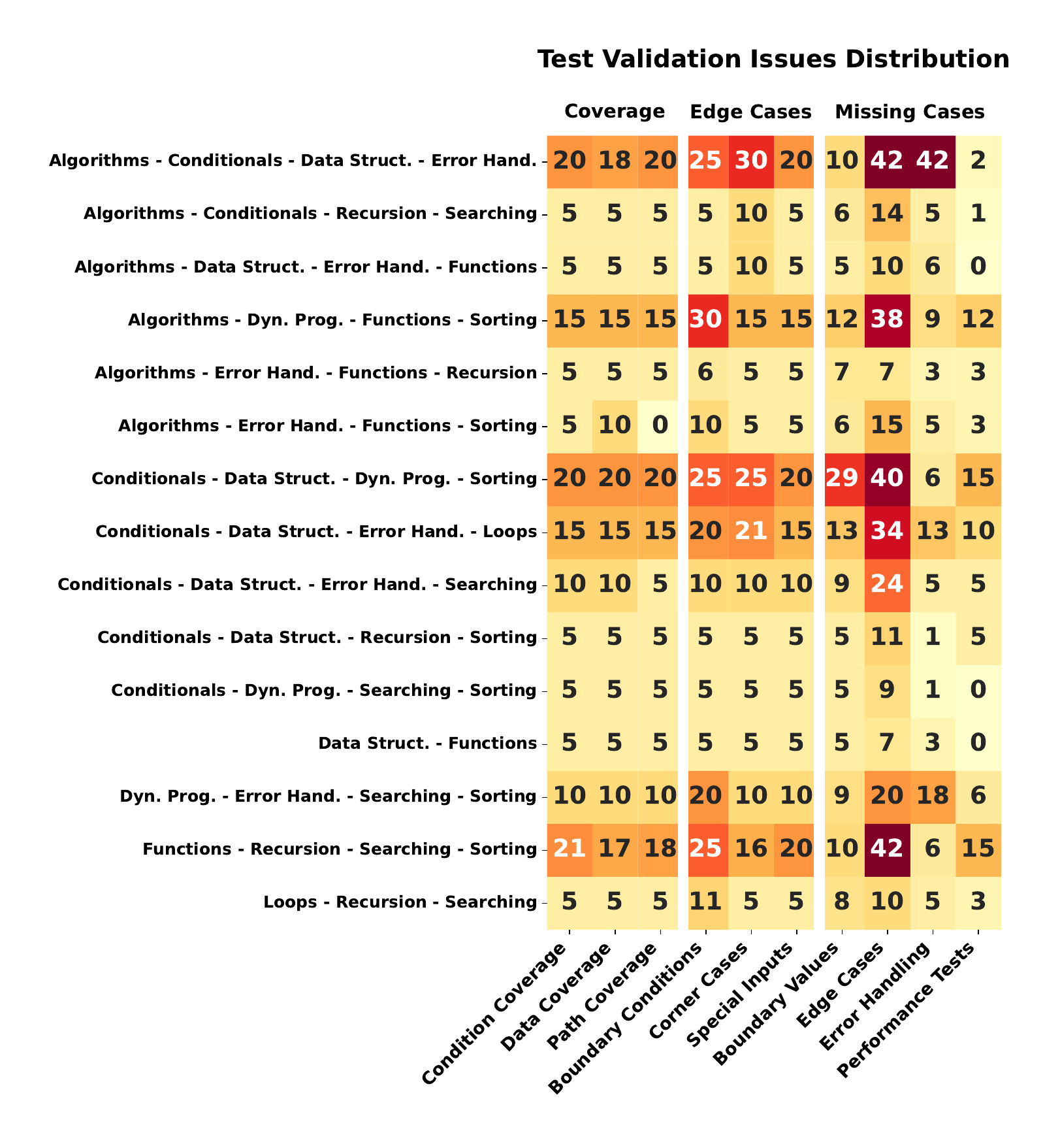}
                \label{fig:4o_test_validation}
            }
            \hfill
            \subfigure[Test validation issues distribution for {\lfourhundredfiveb}]{
                \includegraphics[width=0.45\textwidth]{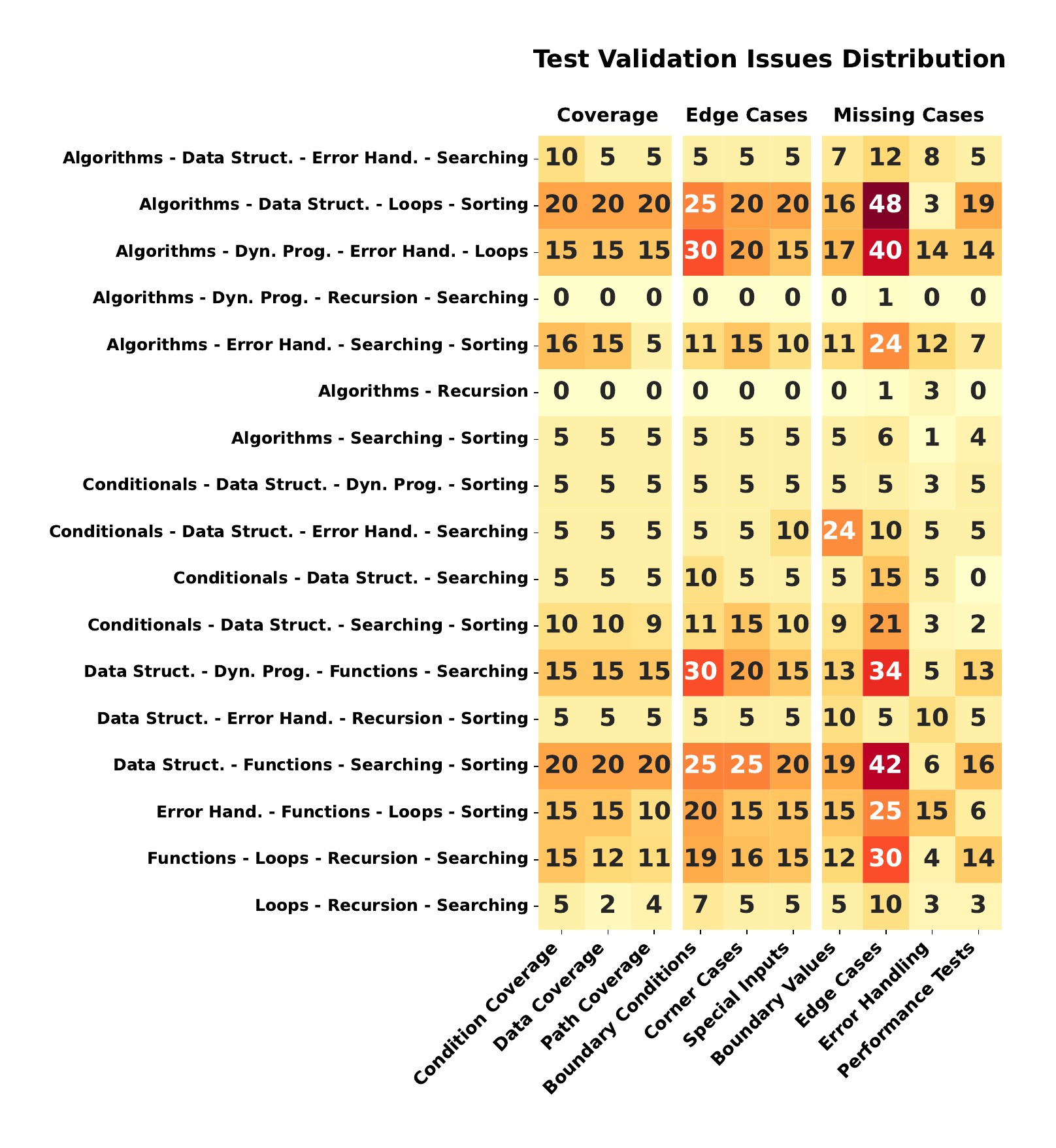}
                \label{fig:l405_test_validation}
            }
            \caption{Test validation issues distribution for {\gfouro} and {\lfourhundredfiveb} averaged over 3 runs. The numbers in each cell indicate the number of times each issue was identified.}
            \label{fig:40_l405_test_validation}
        \end{figure}

        The most encountered error types (algorithm implementation, case sensitivity, and index errors) are consistently related to implementation details rather than fundamental algorithmic understanding. This observation is reinforced by the notably lower frequency of type, setup, and corner case errors across both models and all tested programming concepts. These patterns suggest that while both models demonstrate sound algorithmic understanding, their primary struggle lie in generating the correct code for solutions and tests. Analyzing test validation issues allows us to pinpoint whether the errors stem from incorrectly generated solutions or incorrectly generated tests by the models. Figure~\ref{fig:40_l405_test_validation} presents the distribution of test validation issues for both {\gfouro} and {\lfourhundredfiveb} across concept combinations. Each cell indicates how often a specific validation problem for a test, such as incorrect condition coverage or incorrect boundary checks, was identified. By comparing these data with the error pattern distributions in Figure~\ref{fig:4o_error_pattern} and \ref{fig:l405_error_pattern}, we can discover correlations between the root cause of encountered errors and the concept areas where those errors were raised most frequently.
        
        \begin{figure}[!htb]
            \centering
            \includegraphics[width=\textwidth]{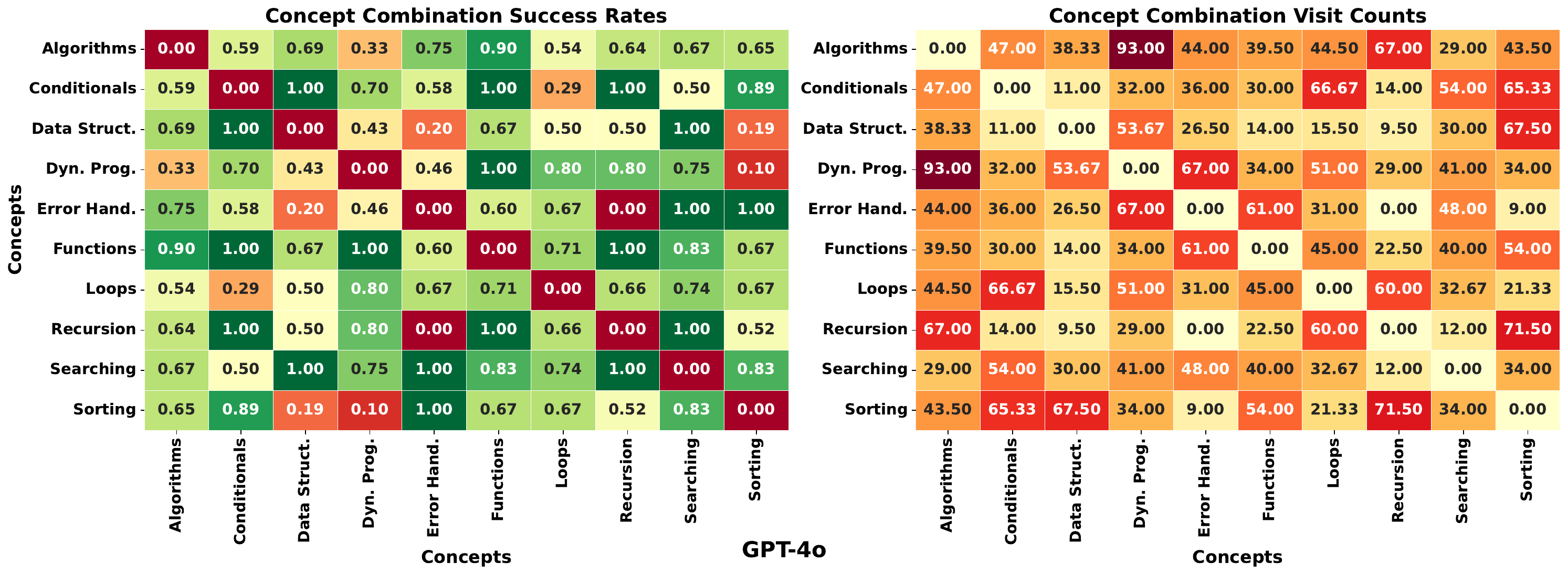}
            \caption {Details on the concept combination effects on {\gfouro}'s performance. The right matrix displays the average success rates for all nodes related to each specific combination. The left matrix displays the average number of times each concept combination was visited in the search tree, regardless of success/failure.}
            \label{fig:4o_concept_heatmap}
        \end{figure}
        For {\gfouro}, the combinations of concepts that have the highest number of test validation failures are [algorithms, conditionals, data structures, error handling] and [functions, recursion, searching, sorting]. These concepts also have the highest number of errors as shown in Figure~\ref{fig:4o_error_pattern}, particularly with index and case sensitivity errors. Furthermore, as reported in Figure~\ref{fig:4o_tree_metrics}, \tool{} has specifically focused on these concepts by generating a high number of nodes for thorough validation and isolation of issues. This indicates that many of {\gfouro}’s generated tests have the same underlying root cause of its generated solutions (for instance, mishandling pointer or array indices). Furthermore, the high frequency of numeric and string value assertions that fail in these tests suggests that {\gfouro} often struggles to produce fully consistent test inputs or expected outputs, leading to assertion failures even when the generated solution is correct. We can observe a correlation between test validation issues and error types, demonstrating that these failures are not only in the generated solutions but also in the generated tests. The highest validation issues appear in concepts requiring numeric and string value assertions. This suggests that {\gfouro} struggles with processing such concepts during test generation. Therefore, even when {\gfouro} produces correct solutions, its limitations in numerical and string processing lead to incorrect test assertions, resulting in failures and error cascades. These issues are also reflected in the success rates and visit counts of concepts as shown in Figure~\ref{fig:4o_concept_heatmap}, where we observe lower success rates and higher visit counts for the combinations of concepts with high error rates and test validation issues. 
        
        \begin{figure}[!htb]
            \centering
            \includegraphics[width=\textwidth]{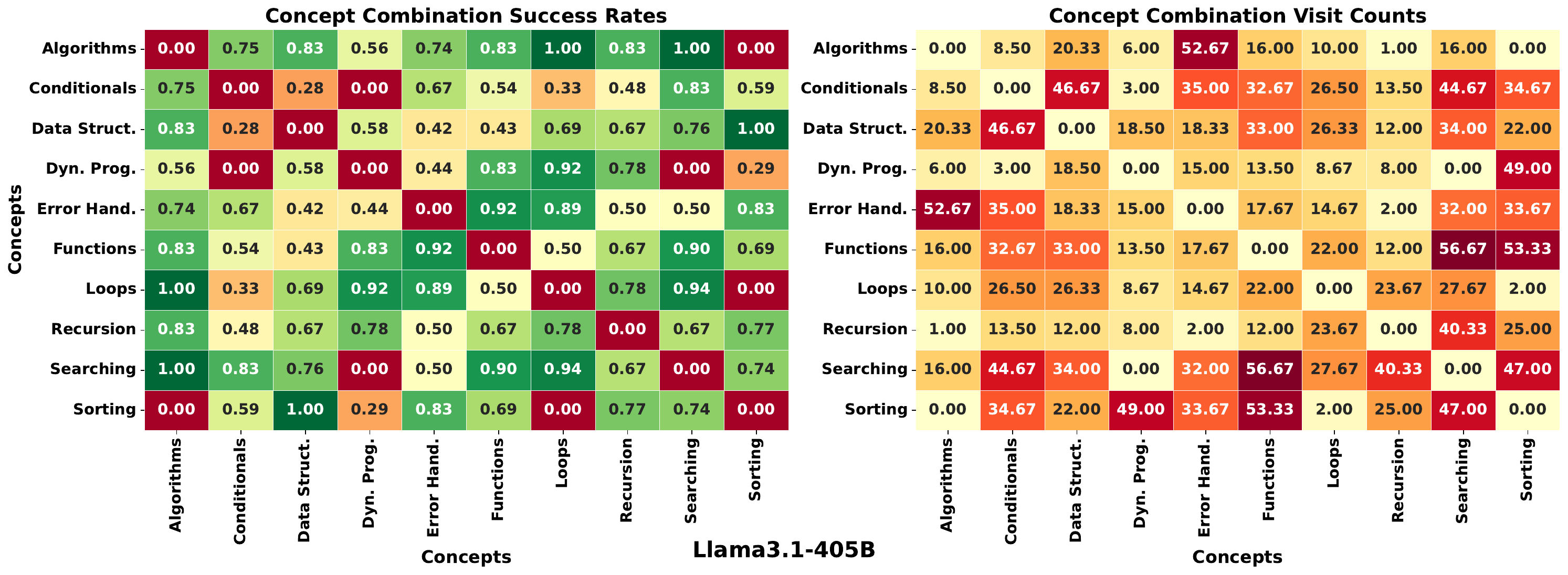}
            \caption {Details on the concept combination effects on {\lfourhundredfiveb}'s performance. The right matrix displays the average success rates for all nodes related to each specific combination. The left matrix displays the average number of times each concept combination was visited in the search tree, regardless of success/failure.}
            \label{fig:l405b_concept_heatmap}
        \end{figure}
\begin{figure}[!htb]
            \centering
            \includegraphics[width=0.5\textwidth]{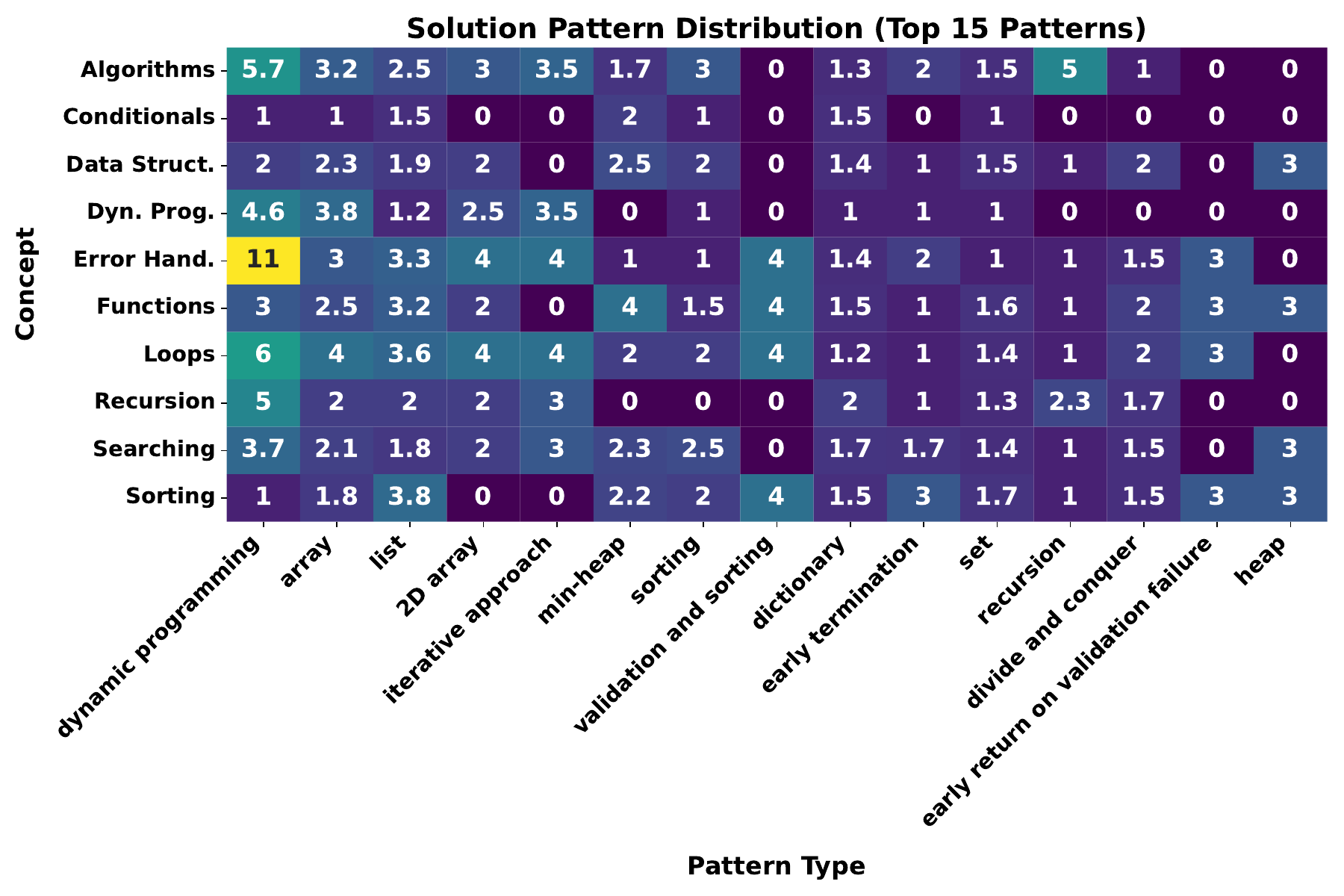}
            \caption{Patterns identified in solutions distribution for {\lfourhundredfiveb} averaged over 3 runs. The numbers in each cell indicate the number of times each pattern was identified.}
            \label{fig:l405_solution_patterns}
        \end{figure}

        {\lfourhundredfiveb} on the other hand, consistently struggles with ``function" and ``sorting" concepts, especially when ``data structures" or ``searching" are also included as shown in Figure~\ref{fig:l405b_concept_heatmap}. The test validation issues in Figure~\ref{fig:l405_test_validation}, demonstrate that combinations like [data structure, function, searching, sorting] exhibit high incorrect coverage issues and a large number of missing or incomplete test cases. Similarly, the error pattern distribution for {\lfourhundredfiveb} in Figure~\ref{fig:l405_error_pattern} shows peaks in case sensitivity and index errors whenever function implementations are tested. We can observe that {\lfourhundredfiveb} exhibits different root causes for failures compared to {\gfouro}, particularly in concepts combination of ``data structures" and ``error handling". By correlating test validation issues and the solution patterns shown in Figure~\ref{fig:l405_solution_patterns}, we can see that {\lfourhundredfiveb}'s failures primarily stem from syntax errors and hallucinations rather than logical errors as evidenced by the high number of failures in using built-in data types (arrays, list, dictionary, etc.). The lowest-performing nodes and their corresponding patterns show that {\lfourhundredfiveb} frequently generates non-existent syntax (e.g., non-existent built-in function calls, incorrect syntax for using built-in data types, etc.) creating a situation where both the generated code and its corresponding tests are incorrect. This leads to the high intervention rates observed in Table~\ref{tab:detailed_concept_performance}, as the \textit{Problem Fixer} repeatedly intervenes to correct both solution and test issues. 

        It is important to note that these insights come from {\tool{}}'s automated analysis. The search trees generated by the framework, enable deeper investigation of behavioral patterns and contain detailed analysis for each node. We only highlight the most significant behavioral patterns observed.
        
    \subsection{Effects of Scale}\label{subsec:appendix:detailed_analysis:effects_of_scale}
        \subsubsection{GPT-4o}
            GPT-4o and GPT-4o Mini, developed by OpenAI, are part of the same model family but differ in scale, performance, and application focus \cite{hurst2024gpt}. GPT-4o is the high-performance, multimodal flagship model optimized for complex tasks requiring deep reasoning and nuanced language understanding while GPT-4o Mini is a lightweight, cost-efficient variant designed for speed and accessibility, prioritizing rapid token generation and affordability. While both models share core architectural features like Transformer-based design and multimodal capabilities, GPT-4o Mini is reported to be significantly smaller than GPT-4o.
            
            \begin{figure}[!htb]
                \centering
                \includegraphics[width=\textwidth]{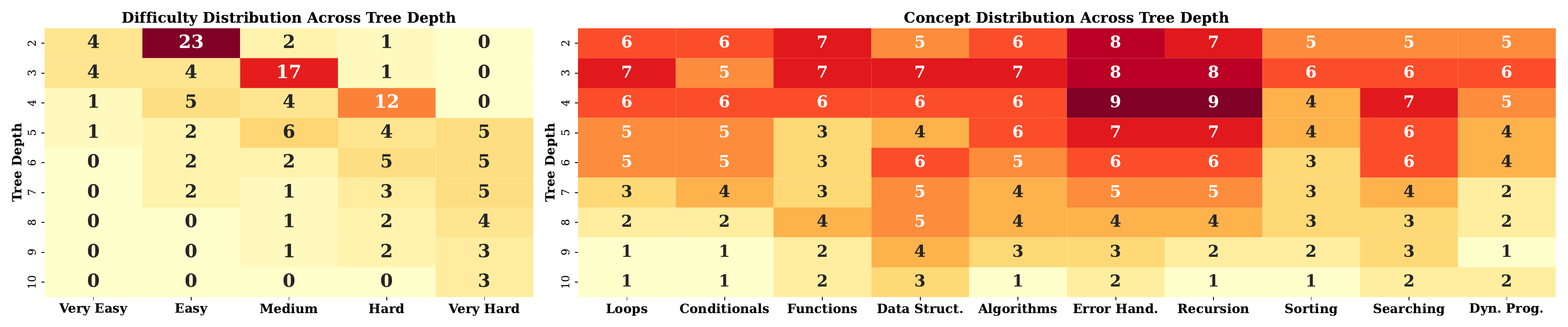}
                \caption{Node distribution for {\gfouromini} averaged over 3 runs. The numbers in each cell indicate the number of nodes.}
                \label{fig:4om_tree_metrics}
            \end{figure}
            Figure~\ref{fig:4om_tree_metrics} displays the node distribution and visit counts of {\gfouromini} throughout the search tree. In the previous sections, we presented performance results for \gfouro, with its corresponding node distributions presented in Figure~\ref{fig:4o_tree_metrics}. Comparing the distributions of {\gfouro} with {\gfouromini} shows us how scale impacts performance. While the majority of {\gfouro}'s nodes are distributed in challenges with ``hard/very hard" difficulty and deeper parts of the tree (as shown in Figure~\ref{fig:4o_tree_metrics}), we can observe that for {\gfouromini}, the majority of nodes are distributed between challenges with ``medium" and ``hard" difficulty and in shallower depths. This is also evident in the search tree for {\gfouromini} as shown in \ref{sec:appendix:trees}.
            
            \begin{figure}[!htb]
                \centering
                \includegraphics[width=\textwidth]{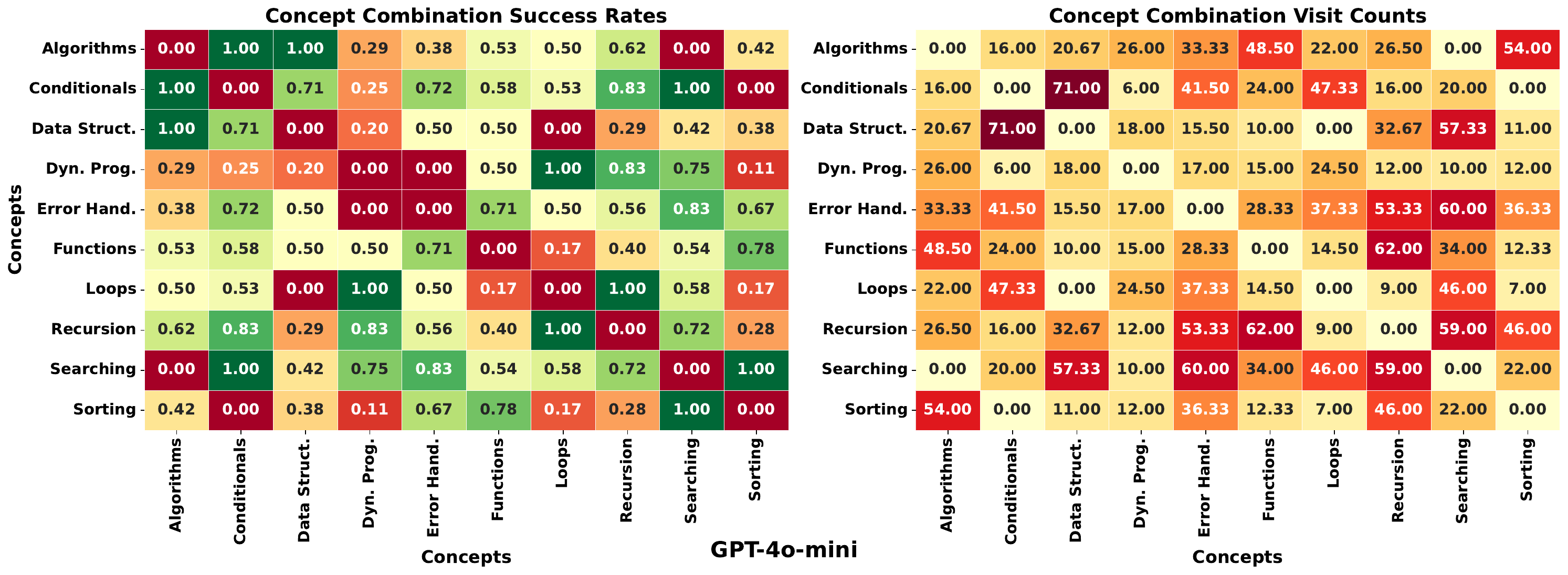}
                \caption{Details on the concept combination effects on {\gfouromini}'s performance. The right matrix displays the average success rates for all nodes related to each specific combination. The left matrix displays the average number of times each concept combination was visited in the search tree, regardless of success/failure.}
                \label{fig:4o_mini_concept_heatmap}
            \end{figure}
            Figure~\ref{fig:4o_mini_concept_heatmap} shows the success rates and visit ratios for nodes corresponding to different concept combinations. As discussed in Section~\ref{sec:experiments}, we can see that the majority of {\gfouromini}'s failures occur when it encounters combinations of concepts that require compositional reasoning. For instance, we can observe that {\gfouromini} has relatively low success rates for ``dynamic programming", which fall even lower when the challenge combines another concept with ``dynamic programming". 

        \subsubsection{Llama 3}
            The Llama 3 herd of models, developed by Meta, is a family of LLMs designed to support multimodality, coding, reasoning, and tool use. The term ``herd of models'' refers to the diverse range of models within the Llama 3 family, each tailored for specific applications \cite{dubey2024llama}. The flagship model, \lfourhundredfiveb{}, is a dense Transformer architecture with 405 billion parameters and a context window of up to 128,000 tokens, enabling it to handle extensive datasets and complex tasks. While these models share foundational training data and post-training processes, they differ in architectural scale—such as the number of layers, model dimensions, attention heads, and FFN dimensions—to optimize performance across varying use cases. This allows us to leverage \tool{} to systematically evaluate how architectural and parametric scale impacts code-generation capabilities. While we have already presented performance results for \lfourhundredfiveb{} (the most capable variant) in prior sections, this section focuses on analyzing performance differences across scaled-down versions of the Llama 3 family.
    
            \begin{figure}[!htb]
                \centering
                    \subfigure[Node distribution for {\lseventyb}]{%
                        \includegraphics[width=0.9\textwidth]{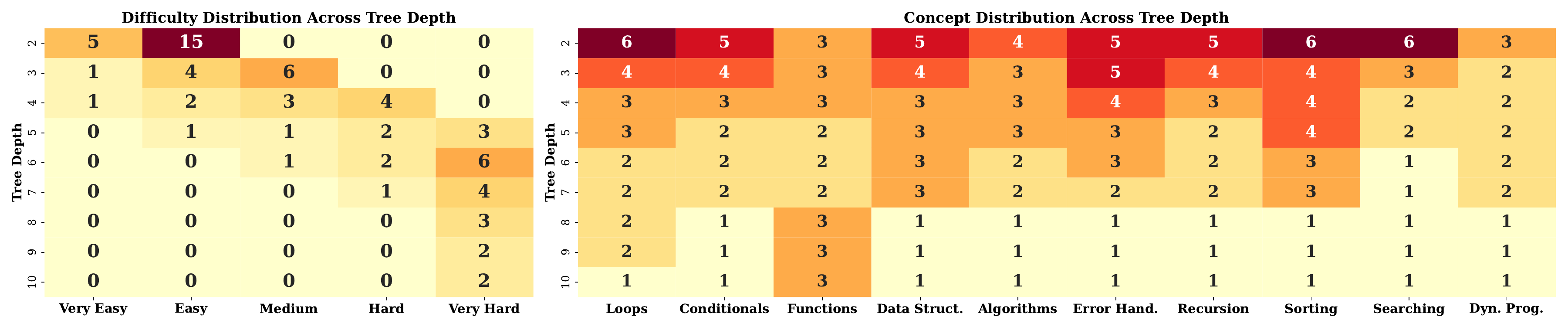}
                        \label{fig:l70_tree_metrics}
                    }\\
                    \subfigure[Node distribution for {\leightb}]{%
                        \includegraphics[width=0.9\textwidth]{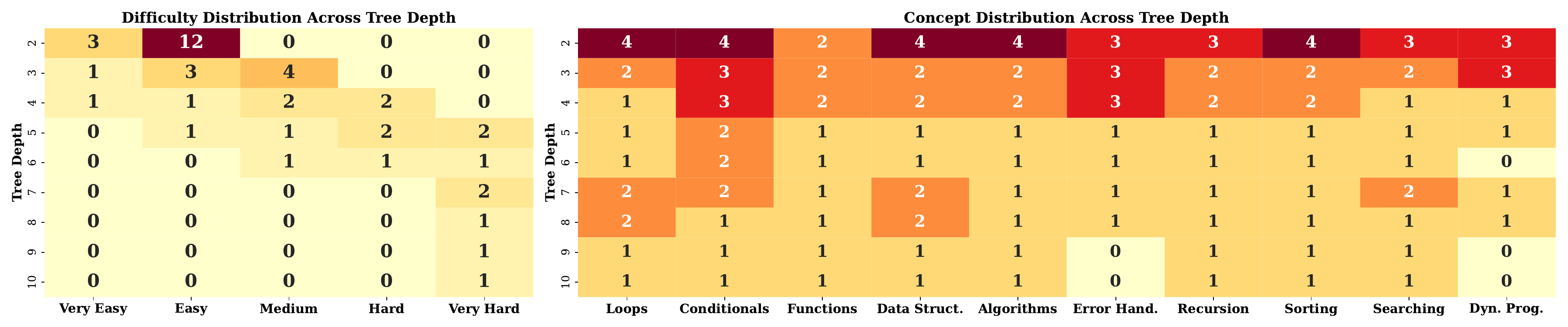}
                        \label{fig:l8_tree_metrics}
                    }
                \caption{Node distributions for {\lseventyb} and {\leightb} averaged over 3 runs. The numbers in each cell indicate the number of nodes.}
                \label{fig:llama_combined_tree_metrics}
            \end{figure}
            
\begin{sidewaysfigure*}[h]
        \centering
        \subfigure[MCTS - Llama3.1-70b]{%
            \includegraphics[width=\textwidth]{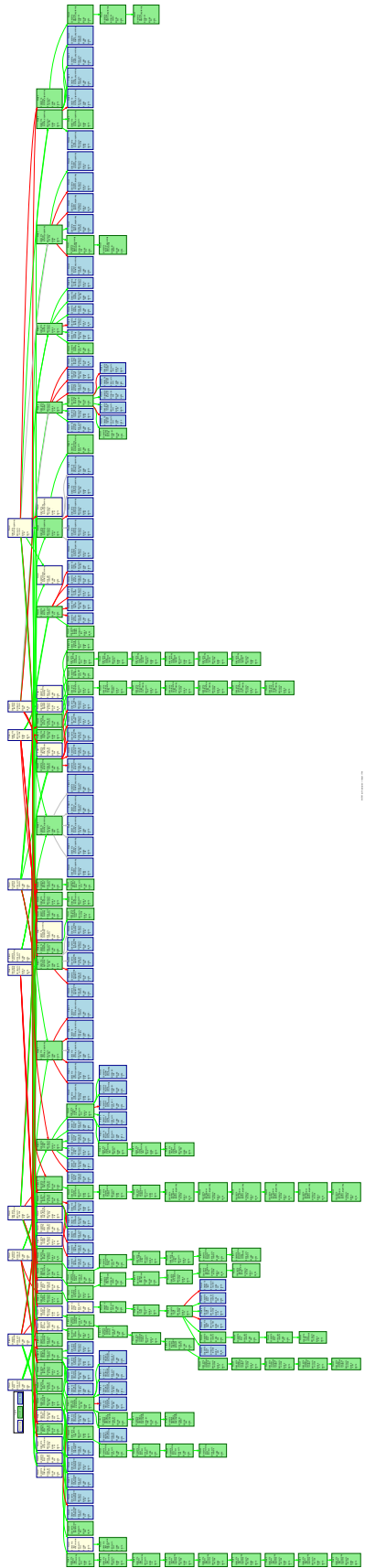}
            \label{fig:tree:mcts_l70b}
        }
        \hfill
        \subfigure[MCTS - Llama3.1-8b]{%
            \includegraphics[width=\textwidth]{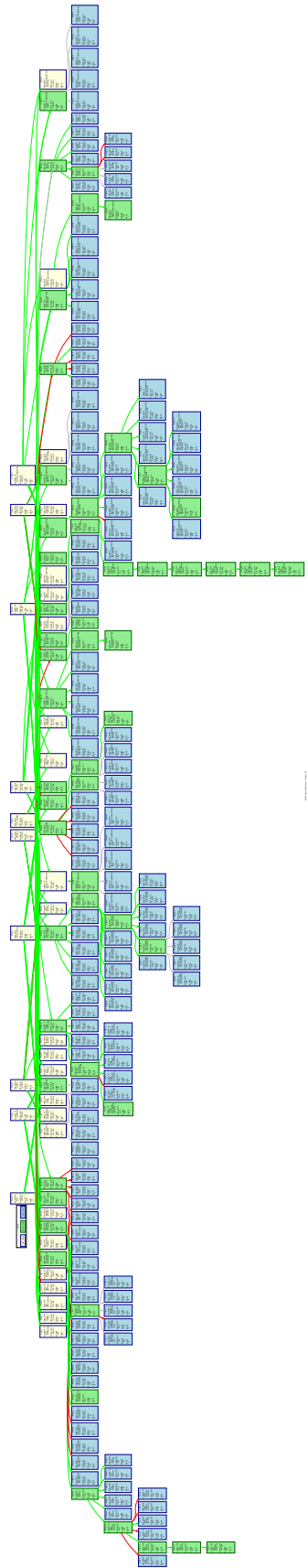}
            \label{fig:tree:mcts_l8b}
        }
        \caption{Search trees generated for {\lseventyb} (a) and {\leightb} (b)}
        \label{fig:tree:mcts_comparison}
    \end{sidewaysfigure*}

            As shown in Figure \ref{fig:l70_tree_metrics} and \ref{fig:l8_tree_metrics}, the depth of explored nodes drops sharply for smaller models, indicating limitations in handling more difficult problems. In particular, the node distribution for \leightb{} shows that only the shallow parts of the tree (depths 2 and 3) and ``easy'' challenges have been explored in depth. The inability to explore deeper nodes suggests that the model fails to generate correct solutions when challenges are more difficult or contain multiple programming concepts. Conversely, \lseventyb{} reaches deeper parts of the tree more often and is capable of reaching nodes with ``very hard'' difficulty to an extent (even though it fails at all of them) as shown in Table \ref{tab:detailed_difficulty_performance}. The node distributions of the search trees generated for {\lfourhundredfiveb}, {\lseventyb}, and {\leightb}) clearly demonstrate how models' scale plays an important role in their problem-solving and code-generation capabilities. The search trees themselves for these two models ({\lseventyb and \leightb}) as presented in Figure~\ref{fig:tree:mcts_comparison}, further show how these models struggle to complete challenges as they become more difficult. Figure~\ref{fig:tree:mcts_l8b} shows how the majority of nodes for {\leightb} are generated in Phase~3 (highlighted in blue). As explained in Section~\ref{subsec:methodology:overview}, Phase~3 is responsible for comprehensively inspecting areas of failure and as such, we can see that {\leightb} consistently fails with high failure rates with the majority of nodes being generated in Phase~3. On the other hand, {\lseventyb} shows a slightly better performance as pictured in Figure~\ref{fig:tree:mcts_l70b}. In the same manner as {\leightb}, the majority of {\lseventyb}'s nodes are generated during Phase~3, However, we can see that unlike {\leightb}, many nodes were also generated in Phase~2, indicating that the model was capable of solving some of these challenges albeit with low success rates.
    
            \begin{figure}[!htb]
              \centering
              \includegraphics[width=\textwidth]{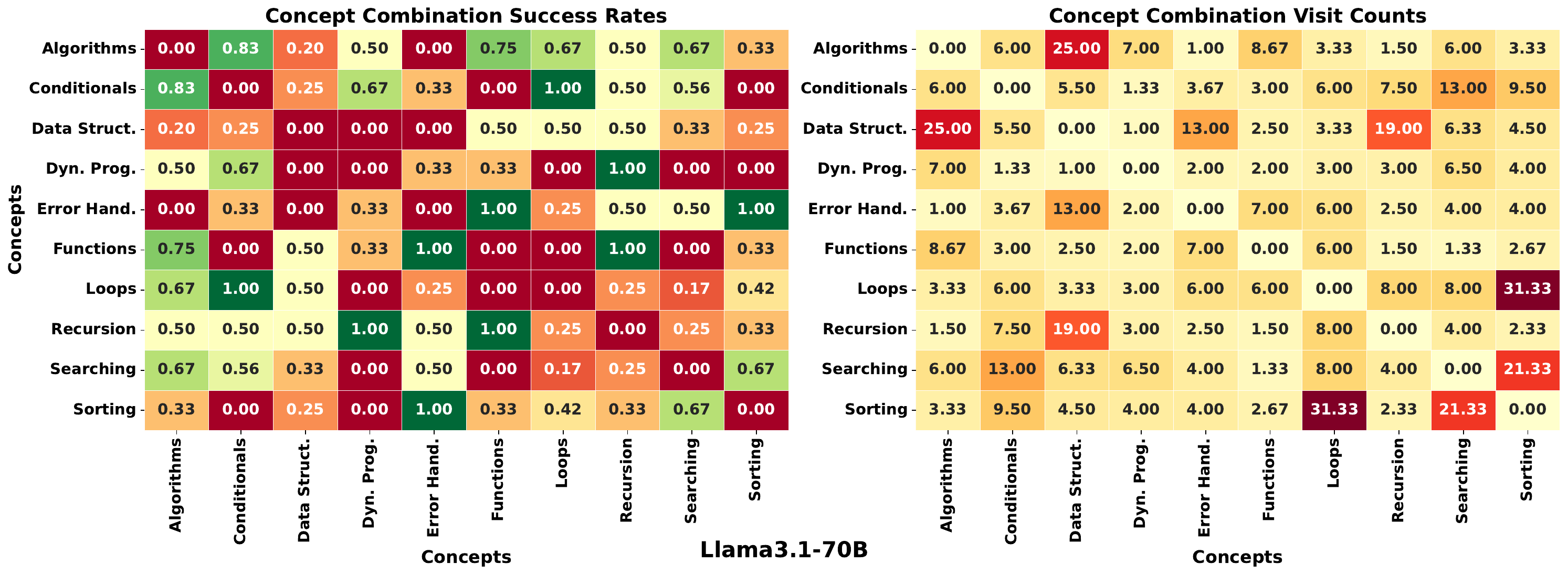}
              \caption {Details on the concept combination effects on {\lseventyb}'s performance. The right matrix displays the average success rates for all nodes related to each specific combination. The left matrix displays the average number of times each concept combination was visited in the search tree, regardless of success/failure.}
              \label{fig:l70b_concept_heatmap}
            \end{figure}
            
            \begin{figure}[!htb]
              \centering
              \includegraphics[width=\textwidth]{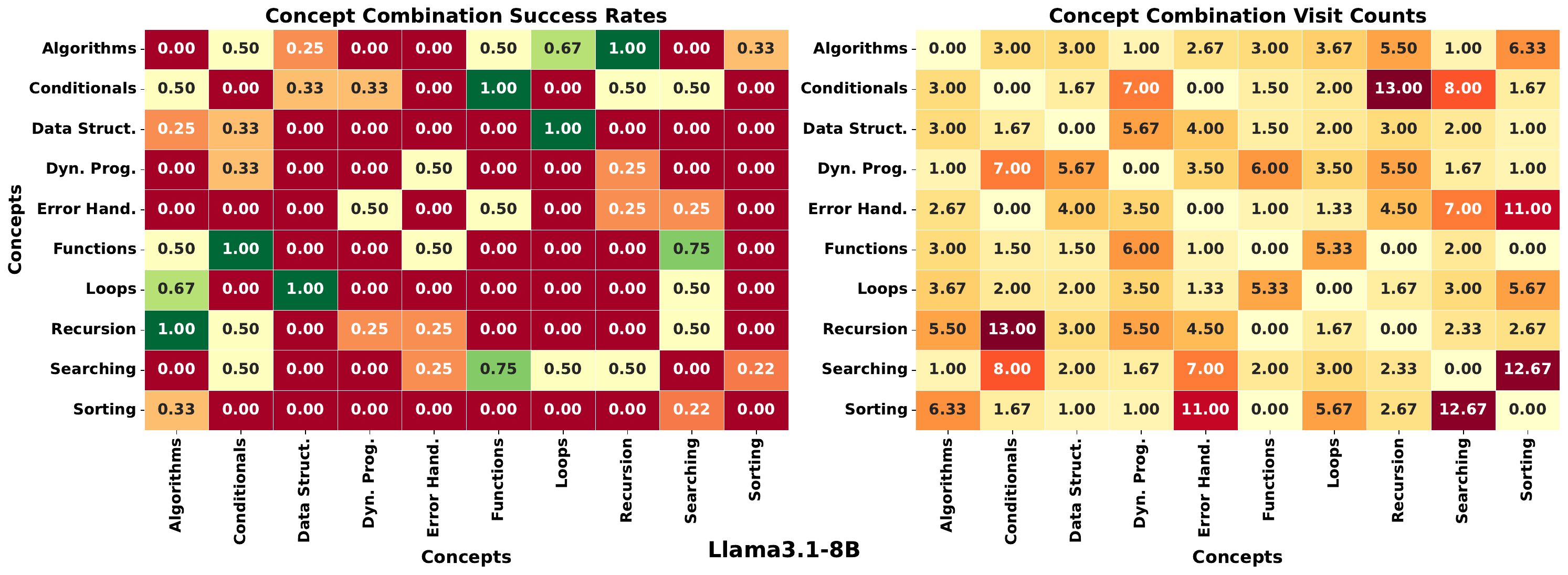}
              \caption {Details on the concept combination effects on {\leightb}'s performance. The right matrix displays the average success rates for all nodes related to each specific combination. The left matrix displays the average number of times each concept combination was visited in the search tree, regardless of success/failure.}
              \label{fig:l8b_concept_heatmap}
            \end{figure}
            Figure \ref{fig:l70b_concept_heatmap} and \ref{fig:l8b_concept_heatmap} further demonstrate performance degradation as the models get smaller. The success rates of concept combinations decrease significantly for \leightb, particularly for tasks requiring more advanced strategies (e.g., ``dynamic programming" or multiple nested constructs). In comparison, \lseventyb{} shows moderate success with simpler ``loops" and ``conditionals" but similarly struggles to sustain the performance under combined, higher-level concepts.

\section{Sample Trees}\label{sec:appendix:trees} 
This section presents the minimized versions of the search trees generated for all models examined in this study. To ensure brevity, the trees presented here focus solely on the concepts, difficulty levels, and scores of each node, along with the phase during which they were generated. The original trees, however, are much more detailed but they would not fit in the content of this paper. The original trees contain the complete challenge description, the generated solutions, tests, attempts, and the corresponding analysis done at each node. This information allows for a comprehensive and fine-grain evaluation of model behavior at each node in the search tree. We have included the full original trees in our replication package \cite{replication_package}. The nodes for each phase are color-coded for distinction, with yellow nodes representing those generated in Phase~1, green nodes representing those generated in Phase~2, and blue nodes representing those generated in Phase~3. The edges between nodes indicate parent-child relationships, with red edges indicating a significant decrease in the child node's TD value compared to its parent, and green edges indicating otherwise. Each node is associated with specific attributes: the concepts related to the node are listed under ``Concepts," the difficulty level is indicated under ``Difficulty," and the number of times the node has been visited is specified under ``Visits."

    \begin{sidewaysfigure*}[h]
      \centering
      \includegraphics[width=\textwidth]{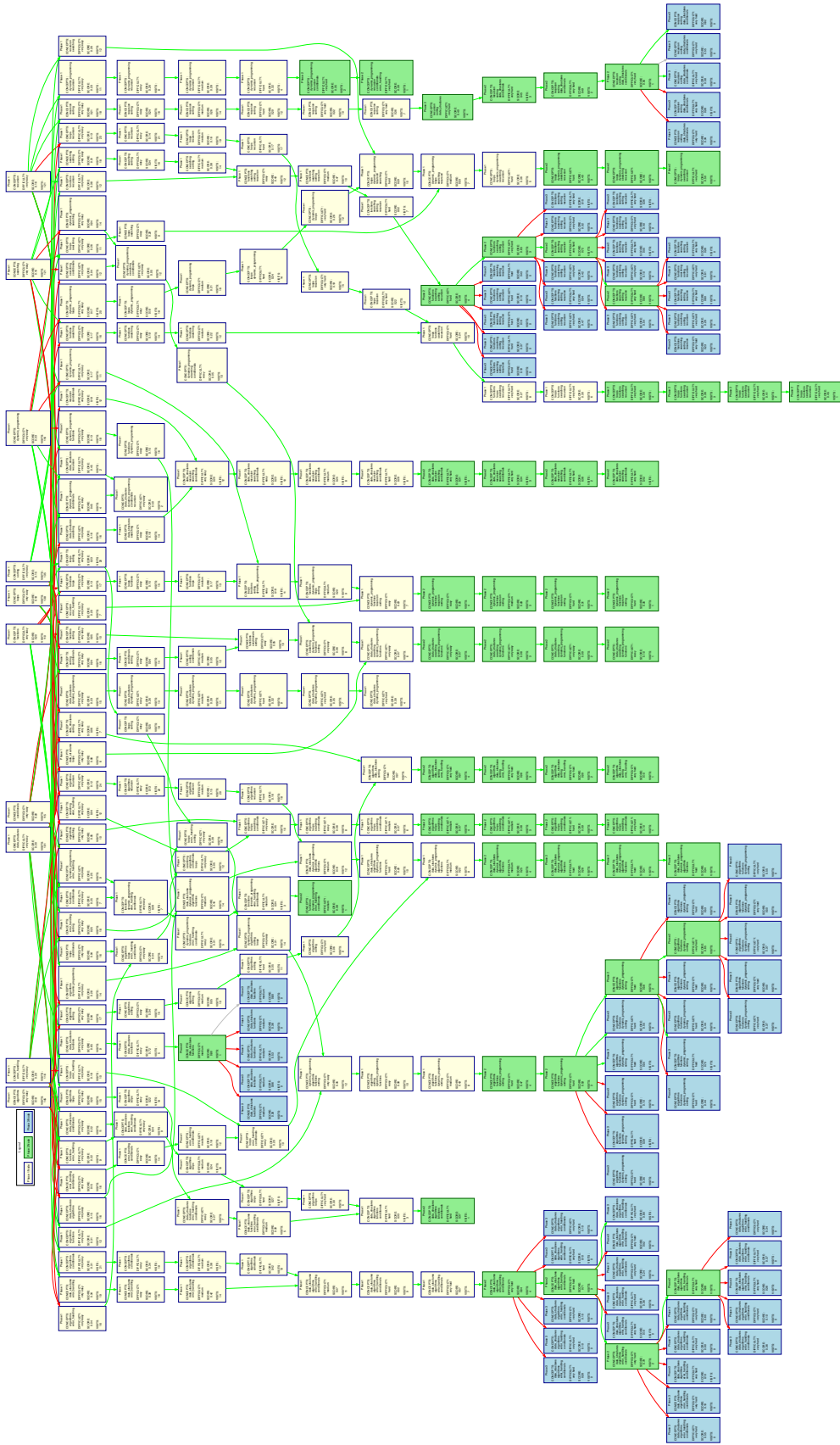}
      \caption{Search tree generated for {\gfouro}}
      \label{fig:tree:mcts_40}
    \end{sidewaysfigure*}
    
    \begin{sidewaysfigure*}[h]
      \centering
      \includegraphics[width=\textwidth]{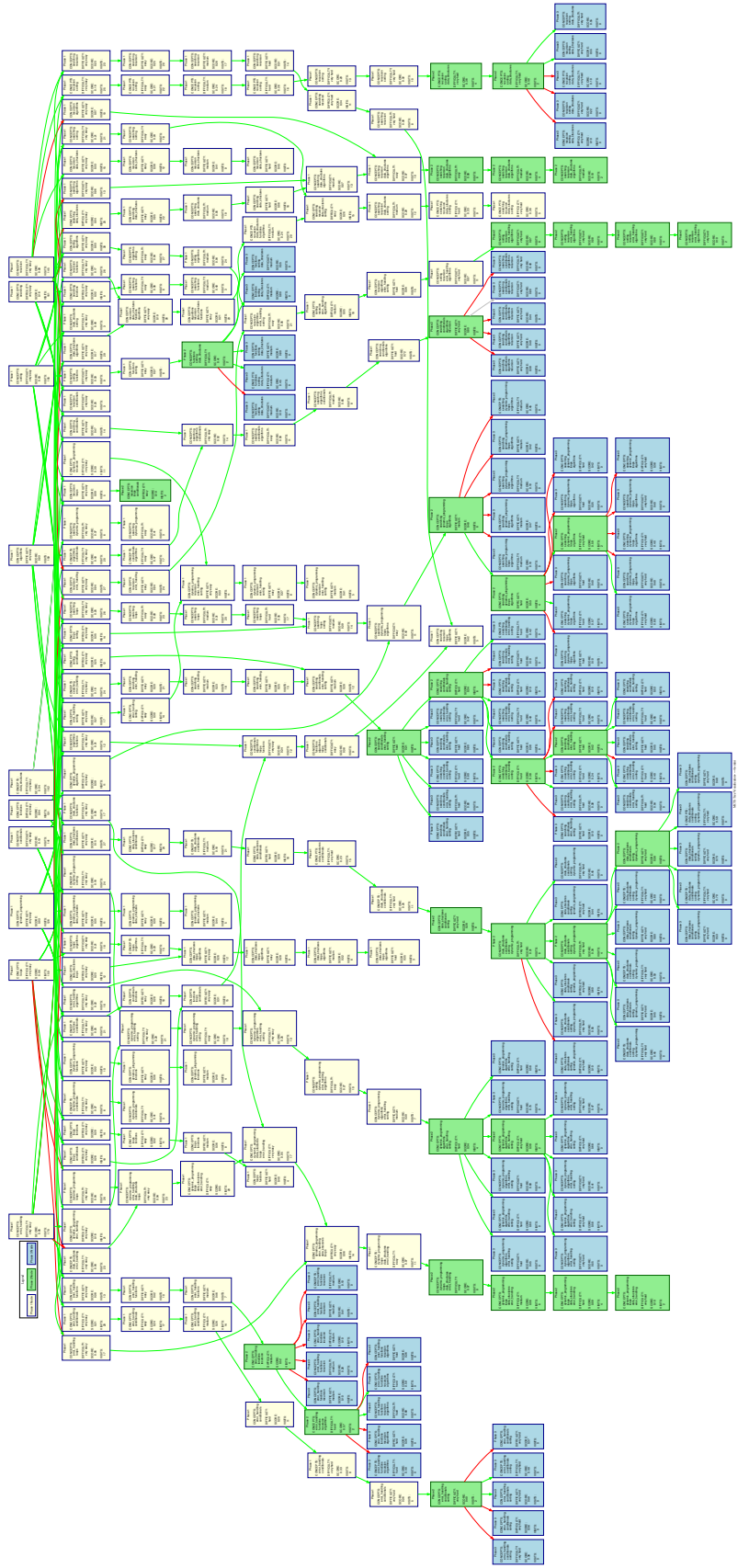}
      \caption{Search tree generated for {\gfouromini}}
      \label{fig:tree:mcts_4o_mini}
    \end{sidewaysfigure*}
    
    \begin{sidewaysfigure*}[h]
      \centering
      \includegraphics[width=\textwidth]{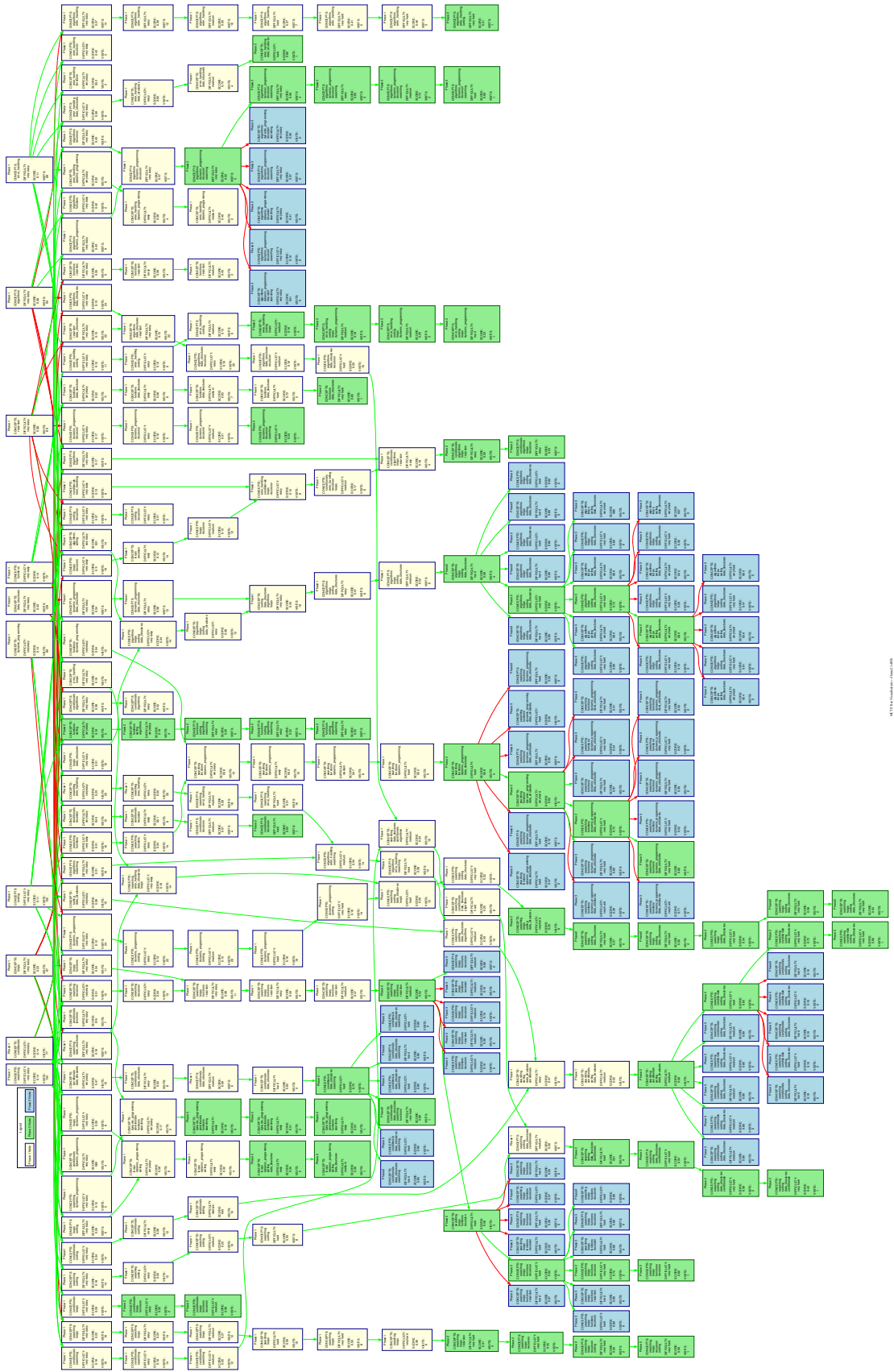}
      \caption{MCTS - Llama3.1-405b}
      \label{fig:tree:mcts_l405b}
    \end{sidewaysfigure*}
    
\end{document}